\documentclass{article}

\usepackage{microtype}
\usepackage{graphicx}
\usepackage{subcaption}
\usepackage{booktabs} 

\usepackage{hyperref}

\usepackage[preprint]{icml2026}

\usepackage{amsmath}
\usepackage{amssymb}
\usepackage{mathtools}
\usepackage{amsthm}
\usepackage{subcaption}

\usepackage[capitalize,noabbrev]{cleveref}

\theoremstyle{plain}

\theoremstyle{definition}

\theoremstyle{remark}

\usepackage[textsize=tiny]{todonotes}

\icmltitlerunning{Beauty and the Beast: Imperceptible Perturbations Against Diffusion-Based Face Swapping via Directional Attribute Editing}

\begin{document}

\twocolumn[
  \icmltitle{Beauty and the Beast: Imperceptible Perturbations Against \\ Diffusion-Based Face Swapping via Directional Attribute Editing}

  \icmlsetsymbol{equal}{*}

  \begin{icmlauthorlist}
    \icmlauthor{Yilong Huang}{sch} and \,  
    \icmlauthor{Songze Li}{sch}
  \end{icmlauthorlist}

  \icmlaffiliation{sch}{Southeast University, Nanjing, China}

  \icmlcorrespondingauthor{Songze Li}{songzeli@seu.edu.cn}

  \icmlkeywords{Face Swapping, Adversarial example, Face editing, Diffusion models}

  \vskip 0.3in
]

\printAffiliationsAndNotice{}  

\begin{abstract}
Diffusion-based face swapping achieves state-of-the-art performance, yet it also exacerbates the potential harm of malicious face swapping to violate portraiture right or undermine personal reputation. This has spurred the development of proactive defense methods. However, existing approaches face a core trade-off: large perturbations distort facial structures, while small ones weaken protection effectiveness. To address these issues, we propose FaceDefense, an enhanced proactive defense framework against diffusion-based face swapping. Our method introduces a new diffusion loss to strengthen the defensive efficacy of adversarial examples, and employs a directional facial attribute editing to restore perturbation-induced distortions, thereby enhancing visual imperceptibility. A two-phase alternating optimization strategy is designed to generate final perturbed face images. Extensive experiments show that FaceDefense significantly outperforms existing methods in both imperceptibility and defense effectiveness, achieving a superior trade-off.
\end{abstract}

\section{Introduction}
\label{sec:intro}

Advancements in AI have enabled increasingly mature face-swapping techniques~\cite{chenSimSwap2020,liFaceShifter2020}, which transfer identity from a source face $I_{src}$ to a target face $I_{tar}$. While valuable in entertainment and media production, the same technology can be maliciously exploited—for instance, to create non-consensual pornographic content or spread disinformation, undermining social trust and even financial markets stability~\cite{zhou2020survey,kaliyar2021deepfake,guo2020future}.

Face swapping methods have evolved from GANs~\cite{goodfellow2020generative}, AEs~\cite{bank2023autoencoders}, and VAEs~\cite{kingma2013auto} to the diffusion models (DMs)~\cite{ho2020denoising,song2020score,song2020denoising}. DMs achieve state-of-the-art realism and identity fidelity~\cite{baliah2024,kimFace2025,zhaoDiffSwap2023a,han2024face}, yet this very realism reduces visual distinguishability, complicating defense against malicious face swapping and enabling more potent disinformation and privacy violations.

Current defense measures against the misuse of face swapping technology include passive detection~\cite{li2020face,yanUCF2023,nguyenLAAnet2024,tanNPR2024} and proactive defense~\cite{ruizDDAdv2020a,aneja2022tafim,wangAntiforgery2022,dongTCA-GAN2023a}. The former detects swapped faces after the fact and cannot completely prevent malicious face swapping. The latter prevents misuse at the source by adding subtle adversarial perturbations to faces, rendering the generated swapped results unusable. This approach can fundamentally prevent the abuse of face swapping, making it a more desirable defense strategy currently.

Proactive defenses are categorized by their perturbation space: pixel, $Lab$ color, and latent space. Pixel-space methods often introduce visible artifacts and $Lab$-space methods distort color tones~\cite{li2025anti}, both harming visual quality. Latent-space methods, such as MyFace~\cite{yam2025my}, avoid pixel-level noise by perturbing abstract latent codes, thereby improving both defense efficacy and imperceptibility. However, latent codes in latent diffusion models (LDMs) are semantically compressed~\cite{rombach2022high}, discarding low-level details (e.g., background, hair color) and retaining only high-level semantic information such as identity ($I_{src}^{id}$). Perturbing this compressed space shifts the latent distribution, leading to decoded facial distortion that worsens with larger perturbations.

\begin{figure}[htbp]
  \vspace{-0.1cm}
    \centering
    \includegraphics[width=0.48\textwidth]{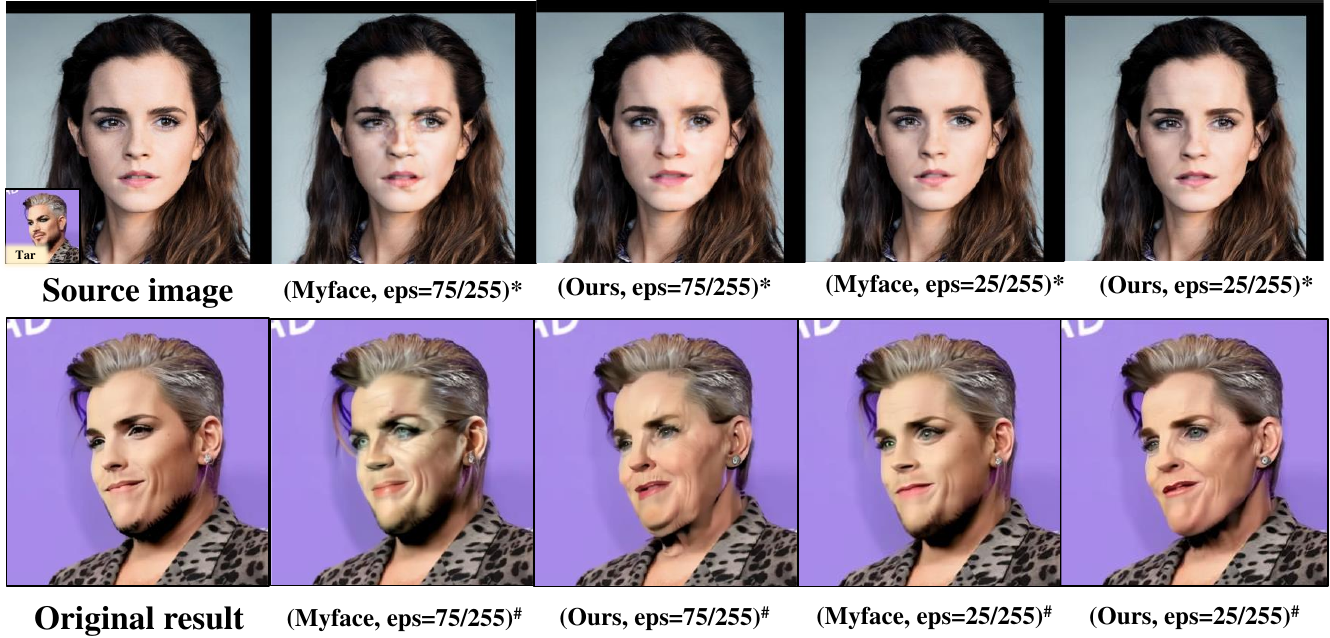}
    \caption{Comparison between the representative method Myface\cite{yam2025my} and our method. “*” denotes adversarial examples, and “\#” denotes face swapping results obtained from adversarial examples. Regardless of the $eps$ value, our method exhibits stronger defense effectiveness and imperceptibility.}
    \label{fig:diff2}
    \vspace{-0.2cm}
\end{figure}

We experimentally confirm this, as shown in Figure~\ref{fig:diff2}. When $eps(\epsilon)=25/255$, the adversarial examples from the Myface with no noticeable facial feature distortion visually, but the defense effectiveness is also very weak. When $eps$ increases to $75/255$, although the defense effect is significant, the method exhibits obvious facial feature distortion (e.g., left eye, left side of the nose), while the compressed parts like hair and background remain almost unaffected. This severely hinders further use of the image by users and can also be detected by adversaries. Consequently, a novel proactive defense method is required to counter advanced DM-based face swapping, one that alleviates facial feature distortion while maintaining defense effectiveness.

To address the issues, We innovatively design a novel adversarial example generation approach to enhance both the defense effectiveness and imperceptibility of it (some examples as shown in Figure~\ref{fig:diff2}). The $I_{src}$ is first subjected to local multi-attribute editing (e.g., \texttt{'wearing\_lipstick'}\allowbreak,\texttt{'mouth\_slightly\_open'}\allowbreak,\texttt{'big\_nose'}\allowbreak, etc.) in the $\mathcal{W}^+$ space\footnote{Proposed by ~\cite{karras2020analyzing}, this space has a good trade-off among image distortion, editability, and perceptual quality}, which is equivalent to introducing an initial perturbation $\delta$. Then, a multi-round iterative optimization is performed to generate adversarial perturbations. Subsequently, the same image undergoes local multi-attribute editing again to restore distorted facial features and artifacts. Finally, the editing process and the adversarial perturbation generation process are optimized in a two-phase alternating manner to balance the defense effectiveness and imperceptibility of adversarial examples. 

Our contributions are summarized in the following aspects:
\begin{itemize}
\item We reveal the underlying cause of facial distortion induced by existing latent-space methods, showing that high-level facial semantic information is more easily susceptible by adversarial perturbations.
\item We propose FaceDefense, a novel method that restores distorted facial features through directional facial attribute editing while preserving adversarial defense effectiveness, thereby concealing the adversarial perturbations.
\item We develop a two-phase alternating optimization strategy to address the min-max optimization problem in attribute editing and adversarial perturbation generation, enhancing both imperceptibility and defense effectiveness of the generated adversarial examples.
\end{itemize}

\section{Preliminaries and Related Work}
\label{sec:background}

This work focuses on defending against malicious DM-based face swapping. Accordingly, this section first introduces the fundamentals of DMs, followed by classic face swapping. We then review representative proactive defense methods in this domain. Finally, we present the facial attribute editing techniques that underpin our approach.

\vspace{-3pt}
\subsection{Diffusion Models}
The core of diffusion models is to approximate the desired original image distribution by progressively denoising a sample from a Gaussian noise distribution. The process consists of two steps: forward noise addition and reverse denoising. Given an original image $\mathbf{x}_0 \sim q(\mathbf{x})$ sampled from the real distribution, Gaussian noise of varying intensity is gradually added to it. Entire forward process is given by:
\begin{equation}
\mathbf{x}_t = \sqrt{\bar{\alpha}_t} \mathbf{x}_0 + \sqrt{1 - \bar{\alpha}_t} \boldsymbol{\epsilon}, \quad \boldsymbol{\epsilon} \sim \mathcal{N}(\mathbf{0}, \mathbf{I}),
\label{eq:forward}
\end{equation}
where $\bar{\alpha}_t = \prod_{s=1}^{t} \alpha_s$,$\quad \alpha_t = 1 - \beta_t$, and $\beta_t$ is the variance of the noise added at step $t$, determined by a predefined schedule of variances $\{\beta_t\}_{t=1}^{T}$.

When $t\to+\infty$, $\mathbf{x}_t$ follows an isotropic Gaussian distribution. To recover $\mathbf{x}_0$, we can train a model $\boldsymbol{\epsilon}_\theta$ using the loss function $\mathcal{L}(\theta) = \mathbb{E}_{\mathbf{x}_0, t, \boldsymbol{\epsilon}} \left[ \| \boldsymbol{\epsilon} - \boldsymbol{\epsilon}_\theta(\mathbf{x}_t, t) \|^2 \right]$  to progressively predict and remove the true added noise $\boldsymbol{\epsilon}$. Following \cite{ho2020denoising}, $\mathbf{x}_0$ can be recovered iteratively via:
\begin{equation}
\mathbf{x}_{t-1} = \frac{1}{\sqrt{\alpha_t}} \left( \mathbf{x}_t - \frac{\beta_t}{\sqrt{1 - \bar{\alpha}_t}} \boldsymbol{\epsilon}_\theta(\mathbf{x}_t, t) \right) + \sigma_t \mathbf{z},
\end{equation}
where $\mathbf{z} \sim \mathcal{N}(\mathbf{0}, \mathbf{I})$. To accelerate sampling, DDIM~\cite{song2020denoising} enables deterministic, skip-step sampling by relaxing the strict Markov assumption while keeping the training objective. Its ODE-form sampling is defined as:
\begin{equation}
\mathbf{x}_{t-1} = \sqrt{\bar{\alpha}_{t-1}} \hat{\mathbf{x}}_0 + \sqrt{1 - \bar{\alpha}_{t-1}} \cdot \boldsymbol{\epsilon}_\theta(\mathbf{x}_t,t).
\label{eq:ddim}
\end{equation}
Here, $\hat{\mathbf{x}}_0 = (\mathbf{x}_t - \sqrt{1 - \bar{\alpha}_t} \boldsymbol{\epsilon}_\theta(\mathbf{x}_t,t)) / \sqrt{\bar{\alpha}_t}$, and $\boldsymbol{\epsilon}_\theta(\mathbf{x}_t,t)$ is the trained noise-prediction model.

\vspace{-3pt}
\subsection{Face Swapping}
Early face swapping relied on hand-crafted feature extraction~\cite{bitouk2008face,lin2012face}. With the rise of deep learning, methods using learned features have become dominant~\cite{chenSimSwap2020,wang2021hififace,cui2023face,li2023e4s}. Such methods typically extract facial features via face recognition models~\cite{deng2019arcface} and integrate 2D/3D prior information~\cite{karras2020analyzing,vaswani2017attention} to preserve identity, yielding highly realistic swapped results. For DM-based face swapping, LDMs are typically adopted~\cite{zhaoDiffSwap2023a,baliah2024,han2024face,kimFace2025} to achieve efficient high-fidelity generation. 
The input $I_{tar}$ is first projected into the latent space via an encoder $\mathcal{E}$, yielding the latent code $\mathbf{z}_0 = \mathcal{E}(I_{tar})$. Then, $\mathbf{z}_t$ is obtained either by Eq.\eqref{eq:forward}, or directly through random Gaussian sampling. Subsequently, a conditional U-Net network $\boldsymbol{\epsilon}_\theta(\mathbf{z}_t, t, \boldsymbol{\tau}_\theta(y))$ is employed to predict the noise that was previously added, where $y$ represents the conditioning information (e.g., $I_{src}^{id}$ and landmarks of $I_{tar}$), and $\boldsymbol{\tau}_\theta$ is a condition encoder (e.g., CLIP or a Transformer). During sampling, classifier-free guidance \cite{ho2021classifier} is applied to enhance conditional control. The noise-prediction term is thereby replaced by:
\begin{equation}
\begin{split}
    \hat{\boldsymbol{\epsilon}}_\theta(\mathbf{z}_t, t, \boldsymbol{\tau}_\theta(y)) = (1+s)\boldsymbol{\epsilon}_\theta(\mathbf{z}_t, t, \boldsymbol{\tau}_\theta(y)) - s\boldsymbol{\epsilon}_\theta(\mathbf{z}_t, t, \varnothing) \nonumber,
\end{split}
\end{equation}
where $s$ denotes the guidance scale and $\varnothing$ indicates an unconditional input. Properly adjusting $s$ allows the swapped face to preserve both diversity and realism. Finally, the denoised latent code $\tilde{\mathbf{z}}_0$ is obtained using Eq.\eqref{eq:ddim} and then decoded back into pixel space via a decoder $\mathcal{D}$, yielding the final face-swapped result $I_{swap} = \mathcal{D}(\tilde{\mathbf{z}}_0)$.

\vspace{-3pt}
\subsection{Face Swapping Proactive Defense}
\label{sec:Proactive}
In the pixel space, Ruiz et al. \cite{ruizDDAdv2020a} employed classical defense methods \cite{goodfellow2014explaining,kurakin2018ifgsm} to successfully disrupt face swapping results generated by GANs. CMUA-Watermark \cite{huang2022cmua} and TAFIM \cite{aneja2022tafim} utilized ensemble attacks on GANs to produce adversarial examples with strong generalizability. TCA-GAN \cite{dongTCA-GAN2023a} targeted FakeApp\cite{fakeapp}, designed a surrogate model and generated high transferability adversarial examples. AdvDM \cite{liangAdversarialExampleDoes2023} was the first to formally define adversarial examples for DMs. Based AdvDM, Mist \cite{liang2023mist} and SDS \cite{xue2023toward} conducted defenses against malicious artistic imitation. DiffusionGuard \cite{choi2024diffusionguard} and PhotoGuard \cite{salmanRaisingCostMalicious2023} designed adversarial examples to defend against malicious image editing. In the $Lab$ color space, Anti-forgery \cite{wangAntiforgery2022} added perturbations to the $a$ and $b$ channels to enhance the imperceptibility of adversarial examples and defend against GAN-based malicious face swapping. In the latent representation space, MyFace applies the PGD method~\cite{madry2017pgd} to defense LDM-based face swapping, effectively preventing malicious face swapping.

\vspace{-3pt}
\subsection{Facial Attribute Editing}
Facial Attribute Editing aims to modify semantic attributes of a face (such as age, gender, and facial features) along given directions to achieve the desired changes and can be categorized into single-attribute editing \cite{lample2017fader,wangHFGANEdit2022,huangSDGAND2024} and multi-attribute editing \cite{choi2018stargan,li2021image,pernuvs2023maskfacegan}. The former maps face images $\mathbf{x}_0$ into the GAN inversion latent space for high-fidelity results, but cannot handle multiple attributes simultaneously. The latter, in contrast, supports concurrent manipulation of multiple attributes. It encodes $\mathbf{x}_0$ into a specific latent space via a dedicated encoder to obtain decoupled style vectors $\mathbf{w}$, which are then optimized for target attributes $\mathbf{a}$ through the mapping $\psi_\mathbf{a}: \mathbf{w} \mapsto \mathbf{w}^*$ under constraints like attribute classification, perceptual, and reconstruction losses. The final high-precision image $\mathbf{x}'_0 = G(\mathbf{w}^*)$ is produced by decoding $\mathbf{w}^*$ with a StyleGAN-based decoder.

\vspace{-5pt}
\section{Problem Formulation}
\label{sec:Method}

\begin{figure*}[t]
    \centering
    \includegraphics[width=0.85\textwidth]{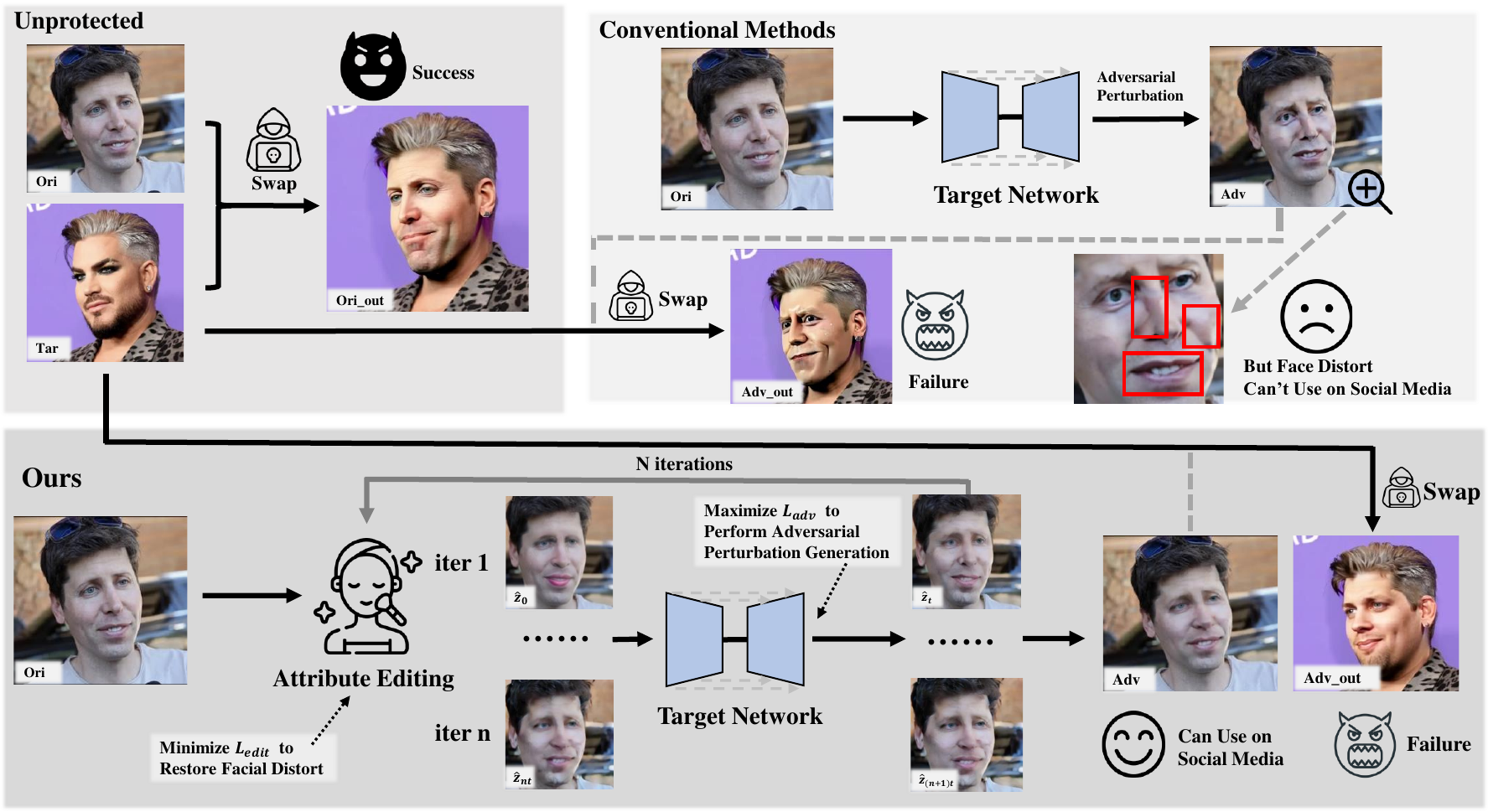}
    \caption{The overall flowchart of FaceDefense, where “Swap” denotes the malicious face-swapping operation, “Target Network” refers to the LDMs that requires defense, “Attribute Editing” indicates the module with directional multi-attribute editing function.}
    \label{fig:flowpic}
    \vspace{-0.4cm}
\end{figure*}

We consider a proactive defense scenario in which a defender (ours) generates adversarial perturbations to protect facial images $I_{src}$ before they are uploaded to social media. If such protected images are later subjected to DM-based malicious face swapping, the attack will fail or be largely suppressed. Our goal is to design a defense that maintains high visual quality of $I_{src}$ while effectively disrupting the swapped output. The threat model is formalized below.

\textbf{Attacker’s goal:} Collect source images $I_{src}$ that the attacker intends to use for malicious face-swapping. Using an open-source DM-based face-swapping model $\mathcal{F}(x)$ and random target images $I_{tar}$ (e.g., photos of a pornographic actor), the attacker performs malicious face-swapping to obtain $I_{swap}=\mathcal{F}(I_{src},I_{tar})$.  

\textbf{Attacker’s capability:} The attacker can access to all open-source face-swapping models and images but cannot obtain non-public source images $I_{src}$ of individual users.  

\textbf{Defender’s goal:} Design an adversarial example $\hat{I}_{src}$ that is visually indistinguishable from $I_{src}$ and thus does not hinder the user from sharing online, while causing the swapped result $\hat{I}_{swap}=\mathcal{F}(\hat{I}_{src},I_{tar})$ to degrade noticeably—e.g., introducing artifacts or losing identity fidelity.

\textbf{Defender’s capability:} The defender can only access open-source LDMs, face recognition models, and the user’s own high-resolution source images, but cannot access the random target images that attackers may use, nor the parameters of any fusion-guidance modules (e.g., ControlNet) that may be employed to improve face-swapping accuracy. This relatively passive setting motivates learning a generic adversarial perturbation that remains effective across varying face-swapping models.

\vspace{-5pt}
\section{Method}
We propose FaceDefense, a defense mechanism that proactively protect the $I_{src}$ against malicious face-swapping, without causing perceivable quality degradation. As illustrated in Figure~\ref{fig:flowpic}, FaceDefense consists of alternative invocations of two key components: “Adversarial Perturbation Generation” and “Attribute Editing”. The former is to generate adversarial perturbations for the target diffusion models, while the latter is to restore the resulting facial distortions.  

\textbf{Adversarial Perturbation Generation.} To achieve the strongest defense effect, we aim to optimize:
\begin{equation}
\max _{\delta} \mathcal{L}_{adv}(\mathcal{D}(z_{src}+\delta)), \quad \text { s.t. } \|\delta\|_{p} \leq \epsilon.
\end{equation}
Here, $z_{src} = \mathcal{E}(I_{src})$ is the latent representation, $\delta$ denotes the adversarial perturbation in the latent space, and $\epsilon$ is $p$-norm perturbation budget. The adversarial example is $\hat{z}_{src} = z_{src} + \delta$. We choose to inject perturbations in the latent space to circumvent the inherent robustness of LDMs and to preserve the imperceptibility of $\hat{I}_{src}$. The overall loss function $\mathcal{L}_{adv}$ comprises three components: an identity loss, a noise loss, and a diffusion loss.

The identity loss $\mathcal{L}_{id}$ aims to maximize the discrepancy between the identity features of $I_{src}$ and those of $\hat{I}_{src}$:
\begin{equation}
    \mathcal{L}_{id}(\hat{z}_{src})=1-\text{cos}(\Phi(I_{src}),\Phi(\mathcal{D}(\hat{z}_{src}))),
\end{equation}
where $\Phi$ is a face recognition model used to extract identity features, and $\mathcal{D}(\hat{z}_{src})$ gets the final adversarial image $\hat{I}_{src}$. 

By disrupting the identity features that typically serve as a conditional signal $y$ in the face-swapping pipeline, $\mathcal{L}_{id}$ prevents $\Phi$ from extracting accurate identity information, thereby obstructing the subsequent swapping process.

The noise loss $\mathcal{L}_{dev}$ is specifically designed for the iterative denoising generation process of LDMs:
\begin{align}
\mathcal{L}^{t}_{dev}(\hat{z}_{src}) &= \| \hat{\boldsymbol{\epsilon}}_\theta({z}_t, t, \boldsymbol{\tau}_\theta(\Phi(I_{src}))) \nonumber \\
&- \hat{\boldsymbol{\epsilon}}_\theta({z}_t, t, \boldsymbol{\tau}_\theta(\Phi(\mathcal{D}(\hat{z}_{src}))))\|_2^2, \\
\mathcal{L}_{dev}(\hat{z}_{src})&=\frac{1}{M} \sum_{t=1}^{M}\mathcal{L}^{t_i}_{dev}(\hat{z}_{src}) \quad t_i \xleftarrow{R} \{0, 1, \ldots, T\}. \nonumber
\end{align}
Here, $\mathcal{L}^{t}_{dev}$ denotes the noise loss at timestep $t$. Since the LDM proceeds through timesteps $t = T, T-1, \ldots, 1$ during the reverse denoising process, each timestep has a corresponding predicted noise and thus a separate $\mathcal{L}^{t}_{dev}$. The final noise loss $\mathcal{L}_{dev}$ is obtained by averaging all $\mathcal{L}^{t}_{dev}$ values. To prevent the adversarial example from overfitting to a fixed temporal path and enhances its robustness, we randomly sample timesteps from $\{0, 1, \ldots, T\}$, rather than following the standard sequential order. This encourages the generation result of the adversarial example to deviate from the $I_{src}$ by disrupting the noise-prediction accuracy that is critical for generation quality, thereby interfering with the normal face-swapping flow during denoising.

The two losses are used in most existing methods. To strengthen the defense, we formulate a diffusion loss $\mathcal{L}_{diff}$ based on the LDM denoising principle. Its expression is identical to $\mathcal{L}_{dev}$, but the term $\hat{\boldsymbol{\epsilon}}_\theta({z}_t, t, \boldsymbol{\tau}_\theta(\Phi(I_{src})))$ is replaced by the true noise $\boldsymbol{\epsilon}$ at step $t$, while ${z}_t$ is no longer randomly sampled but is obtained by adding noise to $z_{src}$ according to Eq.\eqref{eq:forward}. $\mathcal{L}_{diff}$ maximizes the $L_2$ distance between the true and predicted noise for the adversarial example, further forcing the face-swapping output to deviate from the source image $I_{src}$.

Finally, the overall adversarial loss is given by $\mathcal{L}_{adv} =\lambda_1\mathcal{L}_{id}+\lambda_2\mathcal{L}_{dev}+\lambda_3\mathcal{L}_{diff}$, where $\lambda_1,\lambda_2,\lambda_3$ are hyper-parameters. 
 
We optimize the adversarial perturbation $\delta$ via the PGD method by iteratively updating it over the facial high-level semantic information, subject to an $\epsilon$-bound. To satisfy the defense requirement while avoiding facial distortion (see Figure~\ref{fig:diff2}), we optimize $\delta$ by aligning it with the high-level semantic distribution of human faces via attribute editing. This essentially “rearranges” $\delta$ so that its effect manifests only as subtle attribute-level modifications. Since $\delta$ perturbs the entire high-level semantic space, simultaneous multi-attribute editing is required for efficient realignment. Consequently, we seek an approach that supports multi-attribute editing, latent-space optimization, strong generalization, and high-resolution directional attribute editing.

\textbf{Attribute Editing.} MaskFaceGAN~\cite{pernuvs2023maskfacegan} is an advanced high-fidelity facial attribute editor built on StyleGAN2~\cite{karras2020analyzing}. It performs editing by optimizing only the latent code, which is then fed—together with random Gaussian noise—into the generator to produce the final result. The method supports multi-attribute editing, aligning well with our requirements. To adapt MaskFaceGAN to the current task, we modify it as follows: first, $\hat{I}_{src}$ is encoded by the e4e~\cite{tov2021designing} model into a multi-dimensional latent vector $\mathbf{w}$ in the $\mathcal{W}^+$ space, i.e., $\mathbf{w}=E(\hat{I}_{src})$. This latent vector is then optimized by minimizing a reconstruction loss, an attribute classification loss, and a size loss to yield the final face-attribute-edited output.

Reconstruction loss $\mathcal{L}_{rec}$ aims to keep the unedited parts of the face unchanged, ensuring the overall facial structure remains consistent before and after editing:
\begin{equation}
    \mathcal{L}_{{rec}} = \|(\hat{I}_{src} - G(\mathbf{w},\boldsymbol{\delta})) \odot \mathbf{M}_{skin}\|^2_2.
\end{equation}
Here, $\odot$ denotes the Hadamard product, $G$ is the StyleGAN2 model, and $\boldsymbol{\delta}$ is a mapping of our generated adversarial perturbation $\delta$ (i.e., $\delta \mapsto \boldsymbol{\delta} \in \mathbb{R}^{\dim(\mathbf{w})}$), which replaces the randomly sampled Gaussian noise to provide stronger guidance for $\mathbf{w}$. $\mathbf{M}_{skin}$ is the skin-region mask of $\hat{I}_{src}$ obtained by a face parsing model $\Psi$. To ensure natural blending, we apply an edge-smoothing operation $\mathcal{B}$ to $\mathbf{M}_{skin}$, denoted as $\mathbf{M}_{skin} = \mathcal{B}\bigl(\Psi(\hat{I}_{src}, \text{skin})\bigr)$. The above loss constrains $\delta$ to remain unchanged in the skin regions—as perturbations there do not cause pixel-level distortions—while allowing modifications in the facial-feature regions, thereby balancing the defense effectiveness and imperceptibility of the adversarial example.

Attribute classification loss $\mathcal{L}_{cls}$ is designed to ensure that the edited attributes conform to the target requirements. Assuming $z$ attributes are to be edited simultaneously:
\begin{equation}
    \mathcal{L}_{cls} = \frac{1}{Z} \sum_{z=1}^{Z} \text{KL}\big(\mathbf{y}^{z}_{true}, C^{z}(G(\mathbf{w},\boldsymbol{\delta}))\big).
\end{equation}
Here, KL denotes the Kullback–Leibler divergence, $C^{z}$ represents the attribute classifier’s prediction for the $z$-th attribute of the edited result, and $\mathbf{y}^{z}_{\text{true}} \in \{ \vartheta, 1-\vartheta \}$ denotes the ground-truth label for the $z$-th attribute, where $\vartheta$ corresponds to the desired editing strength. Since perturbations primarily distort facial features, the selected attributes must collectively cover all key facial regions to ensure comprehensive restoration. $\mathcal{L}_{cls}$ is therefore essential: it guides the edited features to recover a natural appearance while confining changes to the intended attributes.

Size loss $\mathcal{L}_{size}$ is designed to control the area of the edited region: an overly large editing area may impair the overall visual appearance, while an overly small one may fail to manifest the intended editing effect:
\begin{align} 
\mathcal{L}_{size} &= \text{KL}(\varrho \mathcal{S}(\hat{I}_{src}), \mathcal{S}(G(\mathbf{w},\boldsymbol{\delta}))), \\
\mathcal{S}(x) &= \frac{ { \sum_{h,w}}\mathcal{B}( \Psi(x,\mathbf{a})) }{\left |\mathcal{B}( \Psi(x,\mathbf{a}) )\right | }.
\label{eq:12}
\end{align}
Here, $\varrho$ controls the scale of the editing region, $\mathbf{a}$ represents the set of facial regions corresponding to the target attributes, and \(\mathcal{B}\bigl(\Psi(x, \mathbf{a})\bigr)\) is a single-channel mask. The denominator in Eq.\eqref{eq:12} counts the total number of pixels in the mask, while the numerator sums the pixel values along the channel dimension. $\mathcal{L}_{size}$ ensures proper restoration of distorted facial features, preventing unnatural proportions caused by over-editing or under-editing.

The overall editing objective is then defined by minimizing the composite loss $\mathcal{L}_{edit} =\lambda_4\mathcal{L}_{rec}+\lambda_5\mathcal{L}_{cls}+\lambda_6\mathcal{L}_{size}$, where $\lambda_4,\lambda_5,\lambda_6$ are hyper-parameters. This optimization is performed over $\mathbf{w}$ to guarantee precise attribute-editing restoration. After optimization, the final output:
\begin{equation}
    \hat{I}_{src} = \hat{I}_{src}\odot (1-\mathbf{M}_{edit})+G(\mathbf{w},\boldsymbol{\delta}) \odot \mathbf{M}_{edit},
    \label{eq:fusion}
\end{equation}
where $\mathbf{M}_{edit} = \mathcal{B}\bigl(\Psi(\hat{I}_{src}, \mathbf{a})\bigr)$. Notably, $\mathbf{M}_{edit}$ adopts the dynamic update strategy from MaskFaceGAN for precise region localization. Eq.\eqref{eq:fusion} improves the imperceptibility of the adversarial example while retaining its defense efficacy.  

\begin{figure*}[htbp]
    \centering
    \begin{subfigure}{0.24\textwidth}
        \centering
        \includegraphics[width=\textwidth]{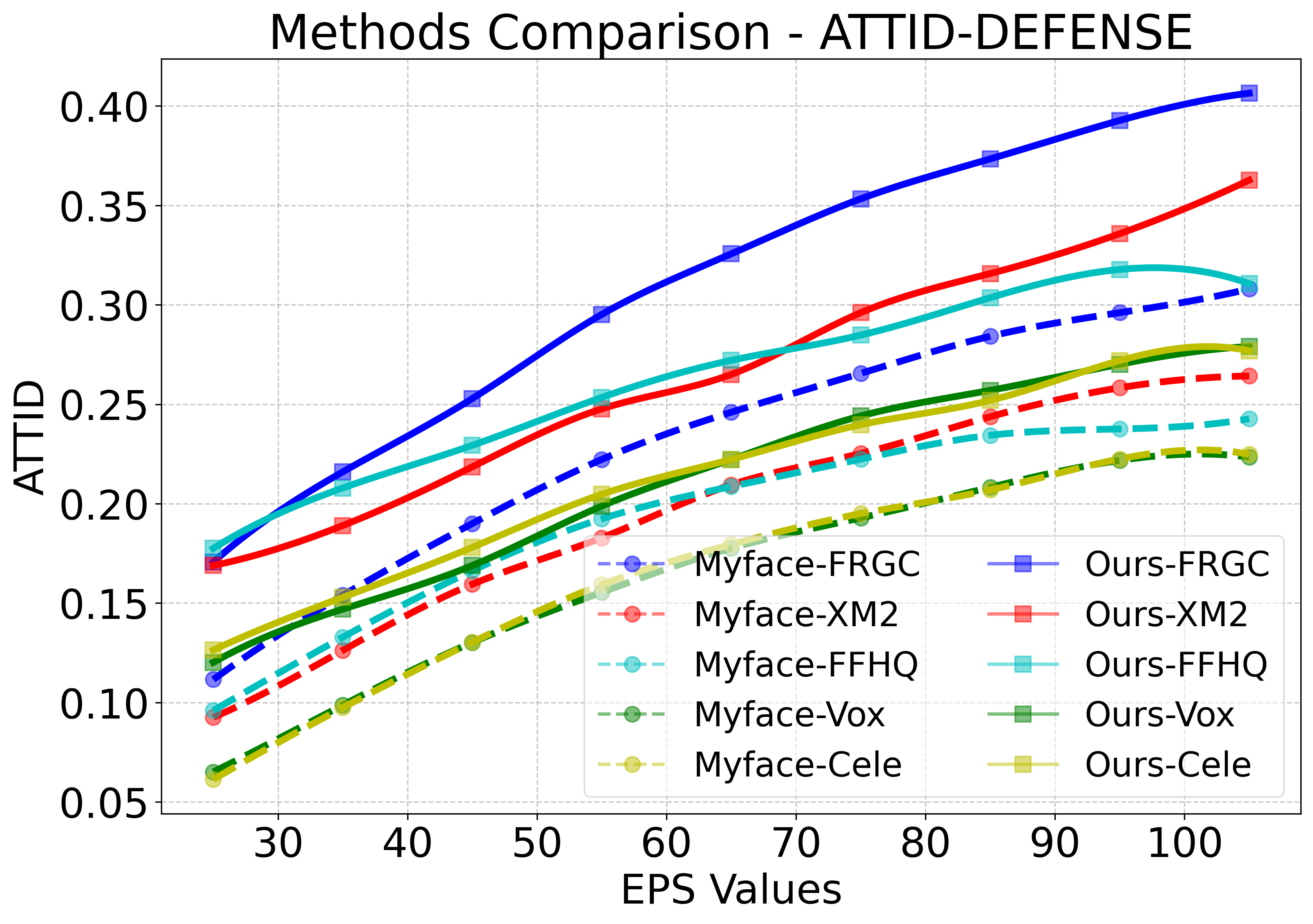}
        \caption{}
        \label{fig:epsATTid}
    \end{subfigure}
    \hfill
    \begin{subfigure}{0.24\textwidth}
        \centering
        \includegraphics[width=\textwidth]{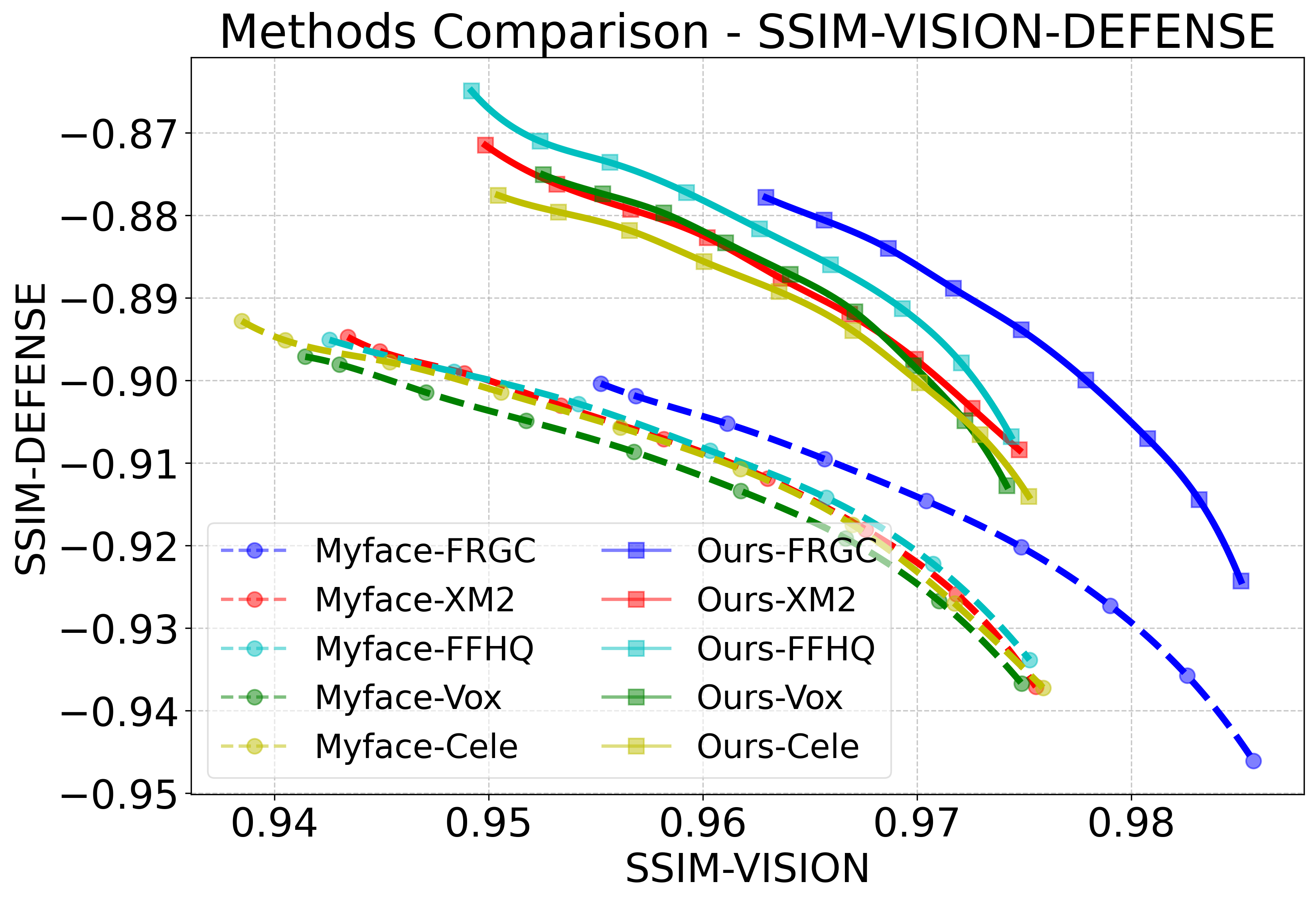}
        \caption{}
        \label{fig:ssimav}
    \end{subfigure}
    \hfill
    \begin{subfigure}{0.24\textwidth}
        \centering
        \includegraphics[width=\textwidth]{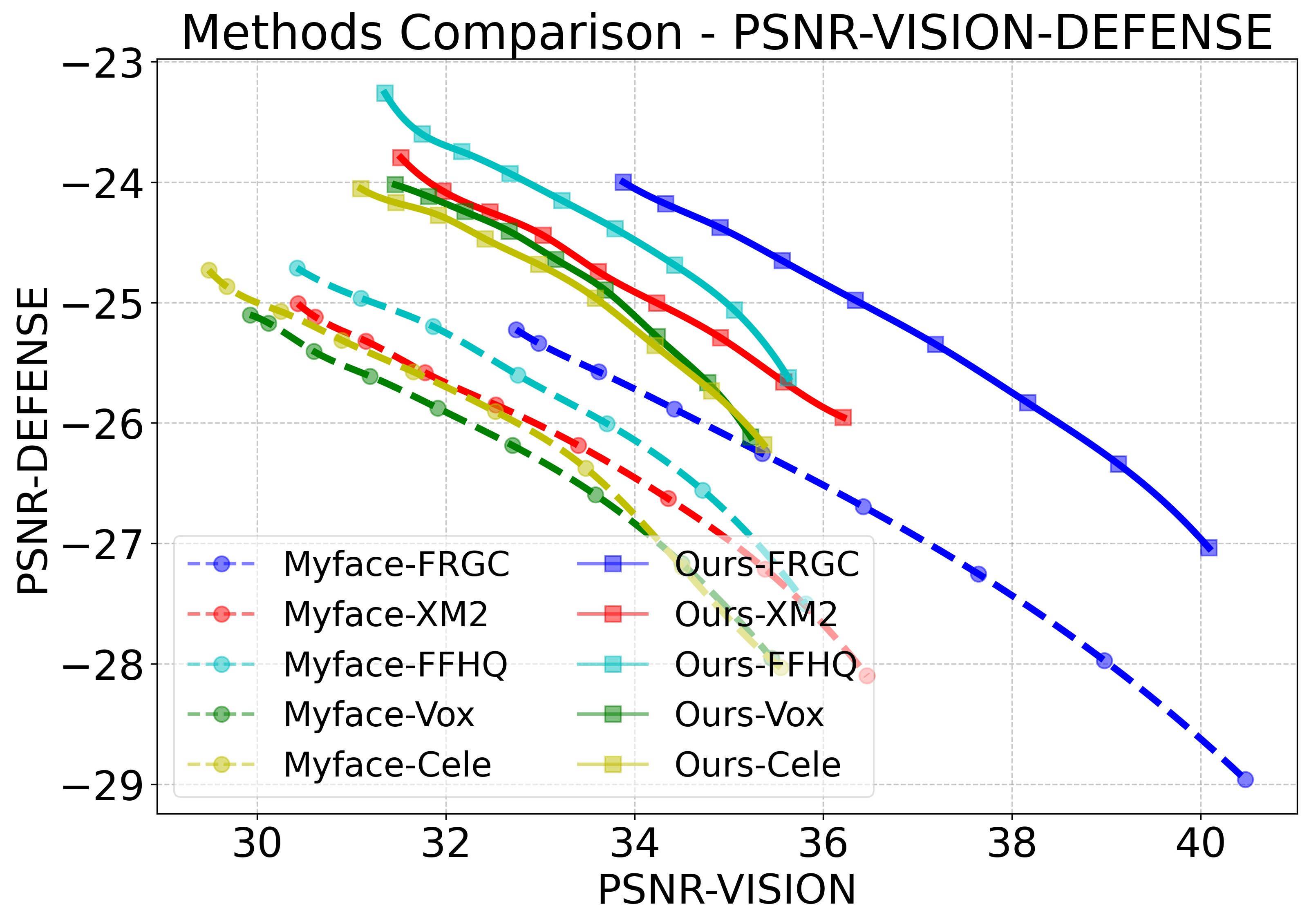}
        \caption{}
        \label{fig:psnrav}
    \end{subfigure}
    \hfill
    \begin{subfigure}{0.24\textwidth}
        \centering
        \includegraphics[width=\textwidth]{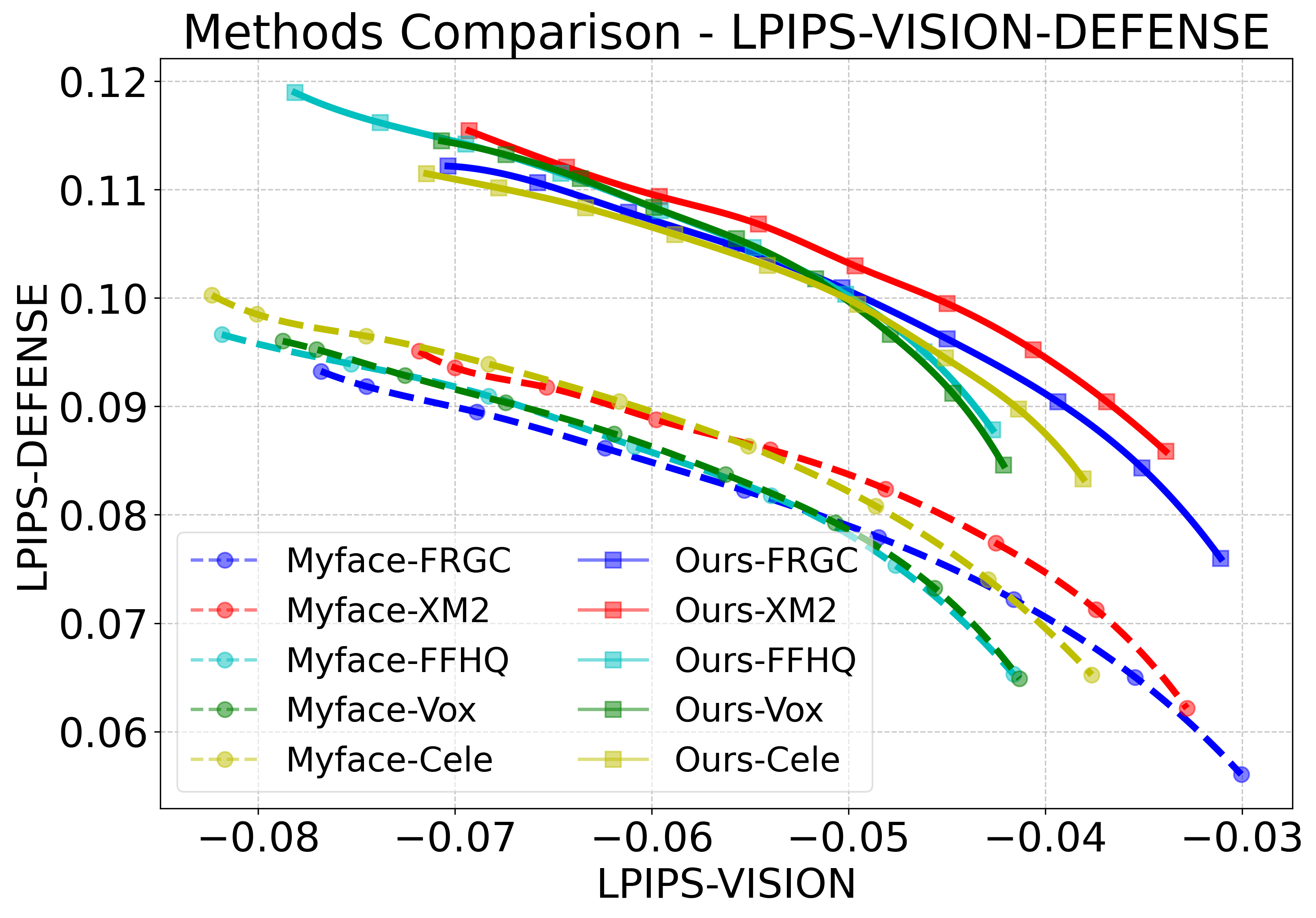}
        \caption{}
        \label{fig:lpipsav}
    \end{subfigure}
    \caption{Evaluation curves across metrics; each marker denotes one experiment. XM2, Vox, and Cele refer to the XM2VTS, VoxCeleb2, and  CelebA-HQ datasets, respectively. All curves are interpolated and smoothed. In (a), curves closer to the top-left corner indicate stronger defense performance. For (b)-(d), metrics favoring lower values have been negated (higher is better) on the y-axes of (b) and (c), and the x-axis of (d), so curves near the top-right corner representing the best trade-off between visual quality and defense efficacy.}
    \label{fig:total_curve}
    \vspace{-0.4cm}
\end{figure*}

\textbf{Two-Phase Alternating Optimization.} To obtain the adversarial example, we need to optimize both $\mathcal{L}_{adv}$ and $\mathcal{L}_{edit}$:
\begin{align}
    \min_{\mathbf{w}} \max_{\hat{z}_{src}} V(\mathbf{w}, \hat{z}_{src}) =  \mathcal{L}_{edit}(\mathbf{w}) + \mathcal{L}_{adv}(\hat{z}_{src}), 
\end{align}
where $\mathbf{w}$ and $\hat{z}_{src}$ can be mutually converted. Jointly optimizing the two losses is straightforward, yet they have opposing goals: $\mathcal{L}_{adv}$ is maximized while $\mathcal{L}_{edit}$ is minimized.  
Simultaneously updating $\mathbf{w}$ and $\hat{z}_{src}$ can cause oscillatory behavior (see Figure~\ref{fig:appzd} in Appendix~\ref{appb:zdsl}). To avoid this, we design a two-phase alternating optimization strategy inspired by GAN training. We first update only $\mathbf{w}$ to edit attributes on the original face, which serves to initialize the adversarial perturbation over the facial high-level semantic information. In the defense phase, we update $\hat{z}_{src}$ to maximize $\mathcal{L}_{adv}$ and then transform it to $\mathbf{w}$. In the restoration phase, we update $\mathbf{w}$ to minimize $\mathcal{L}_{edit}$ and transform it back to $\hat{z}_{src}$. This alternating loop converges to a Nash equilibrium ( see Figure~\ref{fig:appsl} in Appendix~\ref{appb:zdsl}), where the adversarial example simultaneously achieves high defense efficacy and imperceptibility. The complete procedure see in Algorithm~\ref{alg:minimax_alternating}.

\begin{algorithm}[ht!]
  \caption{Two-Phase Alternating Optimization}
  \label{alg:minimax_alternating}
\begin{algorithmic}[1]
\STATE {\bfseries Require:} Source image to protect ${I}_{src}$, target attribute regions $\mathbf{a}$, step sizes $\eta_w$ and $\eta_z$, iteration counts $N,Q,P$, perturbation budget $\epsilon$
\STATE Initialize adversarial example: $\hat{{I}}_{{src}}^{(0)} \gets I_{src}$
\STATE  Initialize perturbation: $\delta^{(0)}=0$
\STATE Initialize multi-dimensional latent vector: $\mathbf{w}^{(0)} \gets E(\hat{{I}}_{{src}}^{(0)})$, $\boldsymbol{\delta} = \boldsymbol{0} \in \mathbb{R}^{\dim(\mathbf{w})}$
\STATE Initialize $\mathbf{M}_{\text{edit}} \gets \mathcal{B}(\Psi(\hat{I}_{src}^{(0)}, \mathbf{a}))$
\FOR{$n = 0$ {\bfseries to} $N-1$}
\STATE \textbf{Optimize} $\mathbf{w}$
    \FOR{$q = 0$ {\bfseries to} $Q-1$} 
        \STATE $\mathbf{g}_{{w}} \gets \nabla_{\mathbf{w}} \mathcal{L}_{{edit}} (\mathbf{w}^{(q)})$
        \STATE Gradient descent update $\mathbf{w}^{(q+1)} \gets \mathbf{w}^{(q)} - \eta_w \mathbf{g}_w$ 
    \ENDFOR
    \STATE Generate editing result ${I}_{{edit}} \gets G(\mathbf{w}^{(q+1)}, \boldsymbol{\delta})$ 
    \STATE $\hat{I}_{{src}}^{(0)} \gets \hat{{I}}_{{src}}^{(0)} \odot (1-\mathbf{M}_{\text{edit}}) + {I}_{{edit}} \odot \mathbf{M}_{\text{edit}}$ 
    \STATE Convert into latent codes $\hat{{z}}_{{src}}^{(0)} \gets \mathcal{E}(\hat{I}_{{src}}^{(0)})$ 
    \STATE \textbf{Optimize} $\hat{z}_{src}$
    \FOR{$p = 0$ {\bfseries to} $P-1$} 
      \STATE ${g}_{{z}} \gets \nabla_{\hat{{z}}_{{src}}} \mathcal{L}_{{adv}}(\hat{{z}}_{{src}}^{(p)})$
      \STATE $\delta^{(p+1)} = \delta^{(p)} + \eta_z \cdot \text{sign}({g}_z)$             
      \STATE $\begin{aligned}
            &\operatorname{Clip} \delta^{(p+1)} \text  { s.t. }  \\
            &\left\|\hat{{z}}_{{src}}^{(p)}+\delta^{(p+1)}-\mathcal{E}({I}_{{src}})\right\|_{\infty} \leq \epsilon
            \end{aligned}$
      \STATE Gradient ascent update $\hat{{z}}_{{src}}^{(p+1)} \gets \hat{{z}}_{{src}}^{(p)} + \delta^{(p+1)}$ 
    \ENDFOR
    \STATE $\hat{{I}}_{{src}}^{(0)} \gets \mathcal{D}(\hat{{z}}_{{src}}^{(p+1)})$      
    \STATE Convert into latent vector $\mathbf{w}^{(0)} \gets E(\hat{{I}}_{{src}}^{(0)})$
    \STATE $\boldsymbol{\delta} \gets \text{map}\big(\delta^{(p+1)}\big) \in \mathbb{R}^{\dim(\mathbf{w}^{0})}$
    \STATE $\mathbf{M}_{\text{edit}} \gets \mathcal{B}(\Psi(\hat{I}_{src}^{(0)}, \mathbf{a}))$
\ENDFOR
\STATE \bfseries{Return:} Adversarial example $\hat{I}_{src}=\mathcal{D}(\hat{z}_{src}^{(p+1)})$ 
\end{algorithmic}
\end{algorithm}

\section{Experiment}
\label{sec:Exp}

This section evaluates the FaceDefense against malicious face swapping. We first compare our method with baseline approaches, followed by ablation studies, robustness tests against image processing operations, and an assessment of adversarial transferability to other models.

\begin{table*}[ht!]
    \caption{Results on the defense effectiveness compared with other methods. The first row indicates the dataset used.}
    \centering
    \scriptsize
    \resizebox{0.98\textwidth}{!}{
    \begin{tabular}
    {c|cccc|cccc|cccc|cccc}
    \toprule
     & \multicolumn{4}{c|}{CelebA-HQ} &\multicolumn{4}{c|}{FFHQ}& \multicolumn{4}{c|}{FRGC} & \multicolumn{4}{c}{XM2VTS}\\
  
\midrule
\textbf{} & SSIM$\downarrow$ & PSNR$\downarrow$ & LPIPS$\uparrow$ & $ATT_{id}\uparrow$  & SSIM$\downarrow$ & PSNR$\downarrow$ & LPIPS$\uparrow$ & $ATT_{id}\uparrow$  & SSIM$\downarrow$ & PSNR$\downarrow$ & LPIPS$\uparrow$ & $ATT_{id}\uparrow$  & SSIM$\downarrow$ & PSNR$\downarrow$ & LPIPS$\uparrow$ & $ATT_{id}\uparrow$\\

\midrule

    AdvDM+   &0.982&31.641&0.036&0.087 &0.956&29.721&0.049&0.093 &\textbf{0.865}&\textbf{23.535}&\textbf{0.116}&0.037 &0.957&30.207&0.043&0.088\\
    AdvDM-   &0.979&34.006&0.025&0.045 &0.969&31.732&0.036&0.053 &0.956&29.929&0.044&0.074 &0.965&31.241&0.036&0.056\\
    DiffusionGuard &0.968&31.838&0.037&0.093  &0.980&33.938&0.024&0.027  &0.962&30.622&0.038&0.064  &0.974&32.832&0.026&0.037\\
    Mist &0.969&32.076&0.036&0.092  &\underline{0.888}&26.019&\underline{0.101}&0.079  &\textbf{0.865}&\underline{23.619}&\textbf{0.116}&0.044  &0.957&30.163&0.044&0.089\\
    PhotoGuard &0.968&31.614&0.036&0.087  &0.952&29.622&0.052&0.107  &0.945&28.790&0.057&0.154  &0.956&30.011&0.046&0.094\\
    SDS+  &0.978&33.792&0.026&0.048  &0.955&29.610&0.049&0.092  &0.950&28.998&0.052&0.118  &0.959&30.174&0.043&0.086\\
    SDS-  &0.967&31.738&0.037&0.096  &0.968&31.651&0.037&0.054  &0.952&29.550&0.047&0.081  &0.964&31.095&0.038&0.057\\
    SDST  &0.987&36.284&0.016&0.026  &0.955&29.934&0.050&0.103  &0.945&28.882&0.056&0.146  &0.957&30.221&0.045&0.087\\
    Myface  &\underline{0.901}&\underline{25.312}&\underline{0.094}&\underline{0.195}  &0.899&\underline{24.623}&0.094&\underline{0.222}  &0.910&25.883&0.086&\underline{0.266}  &\underline{0.903}&\underline{25.582}&\underline{0.089}&\underline{0.225}\\
    \textbf{Ours}  &\textbf{0.886}&\textbf{24.472}&\textbf{0.106}&\textbf{0.240}  &\textbf{0.877}&\textbf{23.929}&\textbf{0.112}&\textbf{0.285}  &\underline{0.889}&24.651&\underline{0.105}&\textbf{0.353}  &\textbf{0.883}&\textbf{24.442}&\textbf{0.107}&\textbf{0.296}\\

      \bottomrule
    \end{tabular}}
    \label{att_supp}
\end{table*}

\begin{table*}[ht!]
    \caption{Results on the imperceptibility compared with other methods. The first row indicates the dataset used.}
    \centering
    \scriptsize
    \resizebox{0.98\textwidth}{!}{
    \begin{tabular}
    {c|ccc|ccc|ccc|ccc}
    \toprule
     & \multicolumn{3}{c|}{CelebA-HQ} &\multicolumn{3}{c|}{FFHQ}& \multicolumn{3}{c|}{FRGC} & \multicolumn{3}{c}{XM2VTS}\\
  
\midrule
\textbf{} & SSIM$\uparrow$ & PSNR$\uparrow$ & LPIPS$\downarrow$  & SSIM$\uparrow$ & PSNR$\uparrow$ & LPIPS$\downarrow$  & SSIM$\uparrow$ & PSNR$\uparrow$ & LPIPS$\downarrow$  & SSIM$\uparrow$ & PSNR$\uparrow$ & LPIPS$\downarrow$\\

\midrule

    AdvDM+   &0.507&18.212&0.411 &0.474&17.600&0.432 &0.289&9.866&0.639 &0.471&19.755&0.425\\
    AdvDM-   &0.653&20.509&0.328  &0.604&19.536&0.353 &0.601&18.588&0.433 &0.599&20.906&0.372\\
    DiffusionGuard  &0.554&18.925&0.413  &0.670&20.361&0.234  &0.660&18.985&0.308  &0.664&21.541&0.249\\
    Mist &0.589&19.244&0.391  &0.383&14.173&0.519  &0.329&10.074&0.649  &0.508&19.867&0.430\\
    PhotoGuard &0.509&18.611&0.403  &0.509&17.819&0.440  &0.543&18.101&0.507  &0.511&19.948&0.426\\
    SDS+  &0.662&20.688&0.335  &0.481&17.981&0.424  &0.476&17.866&0.474  &0.468&19.807&0.417\\
    SDS-  &0.553&18.774&0.411  &0.595&19.236&0.364  &0.609&18.636&0.446  &0.606&20.913&0.382\\
    SDST  &0.730&21.768&0.190  &0.555&18.553&0.414  &0.573&18.459&0.495  &0.541&20.226&0.422\\
    Myface  &\underline{0.924}&\underline{29.306}&\underline{0.101}  &\underline{0.948}&\underline{31.100}&\underline{0.076}  &\underline{0.966}&\underline{34.420}&\underline{0.062}  &\underline{0.953}&\underline{31.781}&\underline{0.060}\\
    \textbf{Ours}  &\textbf{0.960}&\textbf{32.418}&\textbf{0.059}  &\textbf{0.959}&\textbf{32.683}&\textbf{0.065}  &\textbf{0.972}&\textbf{35.564}&\textbf{0.056}  &\textbf{0.960}&\textbf{33.031}&\textbf{0.055}\\

      \bottomrule
    \end{tabular}}
    \label{vis_supp}
    \vspace{-0.4cm}
\end{table*}

\vspace{-3pt}
\subsection{Experimental Settings}
\textbf{Datasets and Metrics.} 
 We evaluate our method on five public high-resolution face datasets: FFHQ~\cite{karras2019style}, CelebA-HQ~\cite{karras2017progressive}, VoxCeleb2~\cite{chung2018voxceleb2}, XM2VTS~\cite{messer1999xm2vtsdb}, and FRGC~\cite{phillips2005overview}. Details are provided in Appendix~\ref{app:Datasets}. For quantitative evaluation, we adopt \textit{defense metrics} and \textit{vision metrics}. The defense metrics compare face-swapping results of the source image and the adversarial example using SSIM, PSNR, and LPIPS~\cite{zhang2018unreasonable}. We also introduce an \textit{identity loss rate} ($ATT_{id}$):
\begin{align}
    ATT_{id} = 1-\frac{\text{cos}(\Phi (I_{src}),\Phi (\hat{I}_{swap}))}{\text{cos}(\Phi (I_{src}),\Phi (I_{swap}))},
\end{align}
where $ATT_{id}$ measures the relative loss rate of identity similarity. A lower $\cos(\Phi(I_{src}), \Phi(\hat{I}_{swap}))$ under stronger defense drives $ATT_{id}$ toward 1, indicating near-complete identity information loss after swapping. For each $I_{src}$, we average $ATT_{id}$ over different $I_{tar}$ as the final score. The vision metrics (imperceptibility) compare $I_{src}$ and $\hat{I}_{src}$ using SSIM, PSNR, and LPIPS—aspects often overlooked in prior work. Further details on parameters configuration are provided in Appendix~\ref{app:para}.

\textbf{Models.} We defend the FaceAdapter face-swapping model~\cite{han2024face} (without access to its ControlNet component). The pre-trained LDM is Stable Diffusion v1-5~\cite{rombach2022high}, and the face recognition model $\Phi$ is ArcFace trained on Glint360K~\cite{deng2019arcface}. The face parser $\Psi$ and attribute classifier $C$ follow MaskFaceGAN~\cite{pernuvs2023maskfacegan}.

\textbf{Baselines.} We compare our method with two categories of baselines. In the latent-space group, we include MyFace~\cite{yam2025my}, which we reimplemented as the official code is not public. In the pixel-space group, we compare against AdvDM+~\cite{liangAdversarialExampleDoes2023}, AdvDM-~\cite{xue2023toward}, Mist~\cite{liang2023mist}, SDS+, SDS-, SDST~\cite{xue2023toward}, DiffusionGuard~\cite{choi2024diffusionguard}, and PhotoGuard~\cite{salmanRaisingCostMalicious2023}.

\begin{table}[t]
\setlength\tabcolsep{6pt}
\caption{Comparison of defense effectiveness on VoxCeleb2.}
\label{main-att}
\vspace{-0.2cm}
\begin{center}
\resizebox{0.42\textwidth}{!}{
\begin{tabular}{ccccc}
\toprule
\multicolumn{1}{l}{}       & \multicolumn{4}{c}{METRIC-DEFENSE}                              \\
VoxCeleb2      & SSIM$\downarrow$ & PSNR$\downarrow$ & LPIPS$\uparrow$ & $ATT_{id}\uparrow$ \\ \midrule
AdvDM+                 & 0.969         & 31.880             & 0.035   & 0.078     \\
AdvDM-          & 0.980         & 34.358             & 0.024   & 0.038     \\
DiffusionGuard  & 0.987        & 36.492             & 0.016  & 0.022             \\
Mist            & 0.969        & 32.157             & 0.036  & 0.090             \\
PhotoGuard      & 0.912        & 28.400             & 0.081  & 0.111              \\
SDS+            & 0.969        & 31.873             & 0.035  & 0.078             \\ 
SDS-            & 0.979        & 34.113             & 0.025  & 0.042             \\ 
SDST            & 0.970        & 32.394             & 0.035  & 0.088             \\ 
Myface          & \underline{0.905}        & \underline{25.610}             & \underline{0.090}  & \underline{0.193}             \\ 
\textbf{Ours}            & \textbf{0.883}        & \textbf{24.407}             & \textbf{0.108}  & \textbf{0.244}             \\ 
\bottomrule
\end{tabular}}
\end{center}
\vskip -0.2in
\end{table}

\begin{table}[t]
\setlength\tabcolsep{14pt}
\caption{Comparison of imperceptibility on VoxCeleb2.}
\label{main-vis}
\vspace{-0.2cm}

\begin{center}
\resizebox{0.42\textwidth}{!}{
\begin{tabular}{cccc}
\toprule
\multicolumn{1}{l}{}       & \multicolumn{3}{c}{METRIC-VISION}                              \\
VoxCeleb2      & SSIM$\uparrow$ & PSNR$\uparrow$ & LPIPS$\downarrow$ \\ \midrule
AdvDM+          & 0.497         & 18.196             & 0.422        \\
AdvDM-          & 0.642         & 20.218             & 0.327        \\
DiffusionGuard  & 0.719        & 21.288             & 0.211            \\
Mist            & 0.547        & 18.757             & 0.426             \\
PhotoGuard      & 0.432        & 14.959             & 0.488              \\
SDS+            & 0.507        & 18.593             & 0.411             \\ 
SDS-            & 0.632        & 19.755             & 0.339             \\ 
SDST            & 0.580        & 19.187             & 0.404            \\ 
Myface          & \underline{0.952}        & \underline{31.196}             & \underline{0.067}             \\ 
\textbf{Ours}            & \textbf{0.961}        & \textbf{32.672}             & \textbf{0.060}             \\ 
\bottomrule
\end{tabular}}
\end{center}
\vskip -0.3in
\end{table}

\subsection{Main Results}
We first compare our method with the latent-space-based approach. To demonstrate the effectiveness of our method across different perturbation budgets, we vary $\epsilon$ within the interval $[25/255, 105/255]$ with a step size of $10/255$. The corresponding VISION-DEFENSE curves and $\epsilon$-$ATT_{id}$ curves for each dataset are illustrated in Figure~\ref{fig:epsATTid}--~\ref{fig:lpipsav}.

Our method consistently outperforms MyFace in $ATT_{id}$ across $\epsilon$ values and datasets. It also achieves higher SSIM, PSNR, and LPIPS scores in both defense efficacy and imperceptibility under nearly all $\epsilon$ settings. At equal defense strength, our adversarial examples are more imperceptible; conversely, under similar imperceptibility, they yield stronger defenses. This allows users to adjust $\epsilon$ to trade off defense effectiveness and imperceptibility. Qualitative comparisons at $\epsilon=75/255$ (following MyFace) on all datasets are provided in Figure~\ref{qunoise-cele}--\ref{qunoise-xm2} (Appendix~\ref{appd:visize_results}).

For pixel-space baselines, we adopt the $\epsilon$ settings from their original papers for fair comparison. Table~\ref{att_supp}--\ref{main-vis} report results on all datasets with MyFace included as reference ($\epsilon=75/255$ for Ours and Myface). For each dataset, the best results are highlighted in \textbf{bold}, and the second-best results are \underline{underlined}. Additional qualitative results across datasets are provided in Figure~\ref{qunoise-cele}--\ref{qunoise-xm2} (Appendix~\ref{appd:visize_results}).

 Overall, compared with pixel-space baselines, our method achieves superior performance in both defense efficacy and imperceptibility, with notably better results in imperceptibility. This confirms its strong capability in defending against malicious face swapping.

\subsection{Ablation Study}
To verify the effectiveness of the introduced diffusion loss $\mathcal{L}_{diff}$ and $\mathcal{L}_{edit}$, we conduct ablation experiments on the CelebA-HQ dataset with $\epsilon = 75/255$. The overall ablation results are summarized in Table~\ref{ablation-att} and Table~\ref{ablation-vis}. Here, “w/o” denotes the removal of a specific component. Best and second-best results are in \textbf{bold} and \underline{underlined}, respectively.

\begin{table}[t]
\setlength\tabcolsep{2.6pt}
\caption{Ablation study on defense effectiveness.}
\label{ablation-att}
\vspace{-0.2cm}
\begin{center}
\resizebox{0.42\textwidth}{!}{ 
\begin{tabular}{lcccc}
\toprule
\multicolumn{1}{l}{}       & \multicolumn{4}{c}{METRIC-DEFENSE}                              \\
CelebA-HQ      & SSIM$\downarrow$ & PSNR$\downarrow$ & LPIPS$\uparrow$ & $ATT_{id}\uparrow$ \\ \midrule
\textbf{Ours}            & 0.886        & 24.472             & 0.106  & 0.240     \\
Ours w/o $\mathcal{L}_{diff}$             & 0.887         & 24.499                & 0.105   & 0.234     \\
Ours w/o $\mathcal{L}_{edit}$       & \textbf{0.872}        & \textbf{23.714}                & \textbf{0.111}  & \textbf{0.267}             \\
Ours w/o $\mathcal{L}_{diff}$,$\mathcal{L}_{edit}$           & \underline{0.875}        & \underline{23.785}             & \underline{0.110}  & \underline{0.260}             \\
\bottomrule
\end{tabular}}
\end{center}
\vspace{-0.3cm}
\end{table}

\begin{table}[t]
\setlength\tabcolsep{7pt}
\caption{Ablation study on imperceptibility.}
\label{ablation-vis}
\vspace{-0.2cm}
\begin{center}
\resizebox{0.42\textwidth}{!}{
\begin{tabular}{lccc}
\toprule
\multicolumn{1}{l}{}       & \multicolumn{3}{c}
{METRIC-VISION}                              \\
CelebA-HQ      & SSIM$\uparrow$ & PSNR$\uparrow$ & LPIPS$\downarrow$ \\ \midrule
\textbf{Ours}            & \textbf{0.960}        & \underline{32.418}             & \underline{0.059}     \\
Ours w/o $\mathcal{L}_{diff}$             & \textbf{0.960}         & \textbf{32.422}                & \textbf{0.058}   \\
Ours w/o $\mathcal{L}_{edit}$       & 0.946        & 30.366               & 0.072             \\
Ours w/o $\mathcal{L}_{diff}$,$\mathcal{L}_{edit}$           & 0.947        & 30.272             & 0.071             \\
\bottomrule
\end{tabular}}
\end{center}
\vskip -0.3in
\end{table}

Removing $\mathcal{L}_{edit}$ yields the strongest defense efficacy but significantly degrades imperceptibility, whereas removing $\mathcal{L}_{diff}$ weakens defense strength while improving imperceptibility. This confirms that both modules shape adversarial-example properties. By combining them, our full method strikes a balance: it sacrifices only marginal defense strength for substantially better imperceptibility, making the adversarial example more practical for real-world use.

\subsection{Robustness Study}
Adversarial examples must maintain effectiveness after common image-processing operations, such as compression on social media or intentional removal attempts by attackers. To simulate these scenarios, we evaluate robustness against Gaussian blur and JPEG compression on the CelebA-HQ dataset with $\epsilon = 75/255$. We compare our method with the robustness-enhanced variant~\cite{yam2025my} of MyFace (denoted as Mdiff). Results are summarized in Table~\ref{robust-att}.

\begin{table}[t]
\setlength\tabcolsep{5pt}
\caption{Robustness evaluation of our method versus Mdiff on CelebA-HQ ($\epsilon=75/255$). “G” denotes Gaussian blur ($\sigma$: intensity) and “J” denotes JPEG compression ($q$: quality).}
\label{robust-att}
\vspace{-0.2cm}
\begin{center}
\resizebox{0.42\textwidth}{!}{ 
\begin{tabular}{lcccc}
\toprule
\multicolumn{1}{l}{}       & \multicolumn{4}{c}{METRIC-DEFENSE}                              \\
CelebA-HQ      & SSIM$\downarrow$ & PSNR$\downarrow$ & LPIPS$\uparrow$ & $ATT_{id}\uparrow$ \\ \midrule
\textbf{Ours}            & 0.886         & 24.472             & 0.106   & 0.240     \\
Mdiff           & 0.920         & 26.541             & 0.078   & 0.184     \\
Ours-G($\sigma=0.5$)  & 0.925        & 26.967             & 0.073  & 0.231             \\
Mdiff-G($\sigma=0.5$) & 0.949        & 29.093             & 0.050  & 0.167             \\
Ours-G($\sigma=1.0$)  & 0.925        & 26.970             & 0.074  & 0.232              \\
Mdiff-G($\sigma=1.0$)    & 0.948        & 29.047             & 0.051  & 0.163             \\ 
Ours-G($\sigma=2.0$)     & 0.926        & 27.113             & 0.075  & 0.233             \\ 
Mdiff-G($\sigma=2.0$)    & 0.947        & 28.952             & 0.054  & 0.161             \\ 
Ours-J($q=30$)  & 0.929        & 27.304             & 0.069  & 0.234             \\
Mdiff-J($q=30$) & 0.949        & 29.169             & 0.050  & 0.178             \\
Ours-J($q=60$)  & 0.926        & 27.028             & 0.072  & 0.231              \\
Mdiff-J($q=60$)    & 0.950        & 29.183             & 0.050  & 0.171             \\ 
Ours-J($q=90$)     & 0.925        & 26.970             & 0.073  & 0.232             \\ 
Mdiff-J($q=90$)    & 0.949        & 29.131             & 0.050  & 0.168             \\ 
\bottomrule
\end{tabular}}
\end{center}
\vspace{-0.4cm}
\end{table}

\begin{table}[t]
\setlength\tabcolsep{6pt}
\caption{Transferability evaluation on FFHQ ($\epsilon=75/255$). Row 1: results on FaceAdapter (defense target); Rows 2–4: results on other models. The best results are highlighted in \textbf{bold}, and the second-best results are \underline{underlined}.}
\label{transfer-att}
\vspace{-0.2cm}
\begin{center}
\resizebox{0.42\textwidth}{!}{
\begin{tabular}{ccccc}
\toprule
\multicolumn{1}{l}{}       & \multicolumn{4}{c}{METRIC-DEFENSE}                              \\
FFHQ      & SSIM$\downarrow$ & PSNR$\downarrow$ & LPIPS$\uparrow$ & $ATT_{id}\uparrow$ \\ \midrule
\textbf{FaceAdapter}            & \textbf{0.877}         & \textbf{23.929}             & \textbf{0.112}   & \underline{0.285}     \\
REFace             & \underline{0.901}         & \underline{24.770}                & \underline{0.085}   & \textbf{0.304}     \\
DiffSwap           & 0.933        & 28.848                & 0.074  & 0.064             \\
DiffFace           & 0.907        & 26.967             & 0.067  & 0.141             \\
\bottomrule
\end{tabular}}
\end{center}
\vskip -0.2in
\end{table}

Our method maintains strong defense efficacy after Gaussian blur and JPEG compression, consistently outperforming Mdiff. This confirms that latent-space perturbations are inherently robust, enabling reliable protection of facial information even after social-media processing. Qualitative results are provided in Figure~\ref{qurobust-guess}--\ref{qurobust-jpeg} (Appendix~\ref{appd:visize_results}).

\subsection{Transferability Study}
Since attackers may use various LDM-based face-swapping models, we evaluate the transferability of our adversarial examples—generated against FaceAdapter—to three other models: DiffSwap~\cite{zhaoDiffSwap2023a}, REFace~\cite{baliah2024}, and DiffFace~\cite{kimFace2025}. Experiments are conducted on the FFHQ dataset with $\epsilon = 75/255$. Quantitative results are shown in Table~\ref{transfer-att}; qualitative comparisons appear in Figure~\ref{trans-reface}--\ref{trans-diffface} (Appendix~\ref{appd:visize_results}).

\begin{figure}[htbp]
    \centering
    \includegraphics[width=0.45\textwidth]{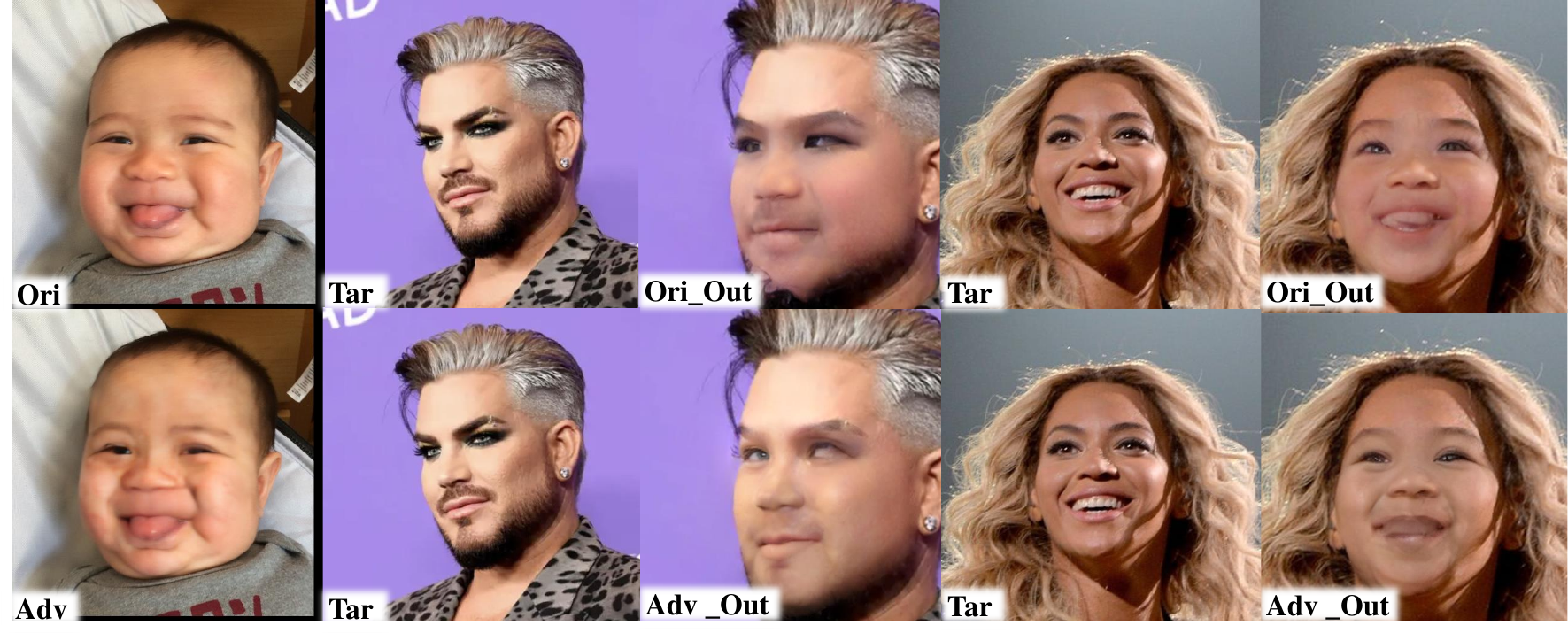}
    \caption{Defense results of our adversarial examples on the DiffSwap~\cite{zhaoDiffSwap2023a} model.}
    \label{fig:diffswap}
    \vskip -0.2in
\end{figure}

Except for the limited effect on DiffSwap (where the reproduced model itself yields low-quality swaps, as shown in Figure~\ref{fig:diffswap}), our adversarial examples maintain noticeable defense effectiveness on other models. Qualitative results show distortions, and alterations in skin tone or gender. This demonstrates strong transferability, offering reliable protection against diverse LDM-based face-swapping models.

\vspace{-3pt}
\section{Conclusion}
\label{sec:clu}

This paper introduces a novel method for defending against DM-based malicious face swapping. The approach integrates attribute editing to restore perturbation-induced distortion and employs a two-phase alternating optimization strategy to balance defense efficacy with imperceptibility. Extensive experiments show that effectiveness of our adversarial examples in protecting user identity information.

\section*{Impact Statement}

The fundamental motivation of this research is to address the direct threats posed by increasingly sophisticated malicious face-swapping techniques to personal privacy, reputation, and social trust. Unlike traditional adversarial attacks that aim to "disrupt" or "deceive" face recognition systems, our method provides individuals with a proactive defense tool. We place particular emphasis on the "imperceptibility" (i.e., user usability) of adversarial perturbations and employ face attribute editing techniques to actively restore facial feature distortions caused by the perturbations. This reflects our commitment to returning control and choice in defense to users, enabling them to protect their facial data from misuse under voluntary circumstances.

However, any adversarial example generation technique inherently carries a dual-use risk. Specifically for our method, there exists a potential risk of reverse engineering. The novel loss function and phased co-optimization strategy designed to enhance defense effectiveness could be analyzed by attackers to improve the robustness of malicious face-swapping models or to inspire the development of new attack methods.

This study uses publicly available face datasets (e.g., CelebA, FFHQ, VoxCeleb2) for training and evaluation. We confirm that these datasets are licensed for research purposes and have taken steps to minimize potential privacy impacts on the involved individuals. Moreover, we disclose the core ideas of the method in detail in this paper, aiming to foster collective scrutiny of such defense paradigms within the security community and to encourage the development of corresponding detection technologies for counterbalance. Finally, we emphasize that this technology must be strictly limited to scenarios where users protect their own facial data. Any application that interferes with others or attacks public systems constitutes malicious misuse of this research, which we explicitly oppose.


\bibliography{example_paper}
\bibliographystyle{icml2026}

\newpage
\appendix

\section{Alternating and Joint Optimization}
\label{appb:zdsl}
\begin{figure}[htbp]
    \centering
    \includegraphics[width=0.46\textwidth]{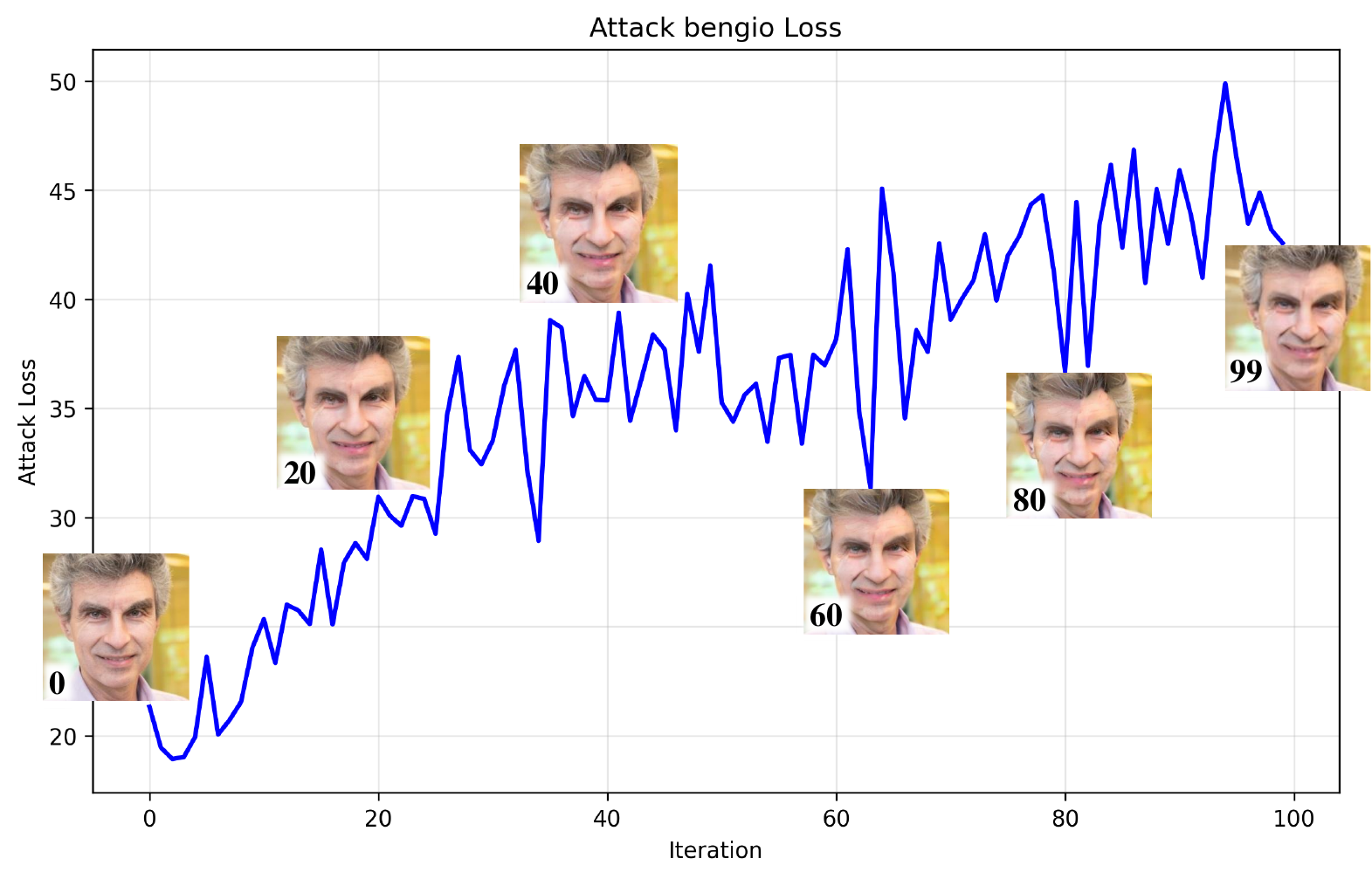}
    \caption{The $\mathcal{L}_{attack}$ curve for the image “bengio.jpg” under alternating optimization. The number on the every image denotes the current iteration round. We extract the adversarial example at current round and convert it to RGB space for visualization. }
    \label{fig:appsl}
\end{figure}

In this section, we demonstrate the loss oscillation caused by jointly optimizing $\mathcal{L}_{dev}$ and $\mathcal{L}_{diff}$, and the convergence of loss functions when using alternating optimization of $\mathcal{L}_{dev}$ and $\mathcal{L}_{diff}$. We set $\epsilon = 75/255$ and use the VoxCeleb2 dataset, and present the values of $\mathcal{L}_{attack} = \mathcal{L}_{adv} - \mathcal{L}_{edit}$ for the source image “bengio.jpg” under different optimization strategies, along with visualizations of intermediate adversarial examples during the iterative process, as shown in Figure~\ref{fig:appzd} and Figure~\ref{fig:appsl}:

\begin{figure}[htbp]
    \centering
    \includegraphics[width=0.46\textwidth]{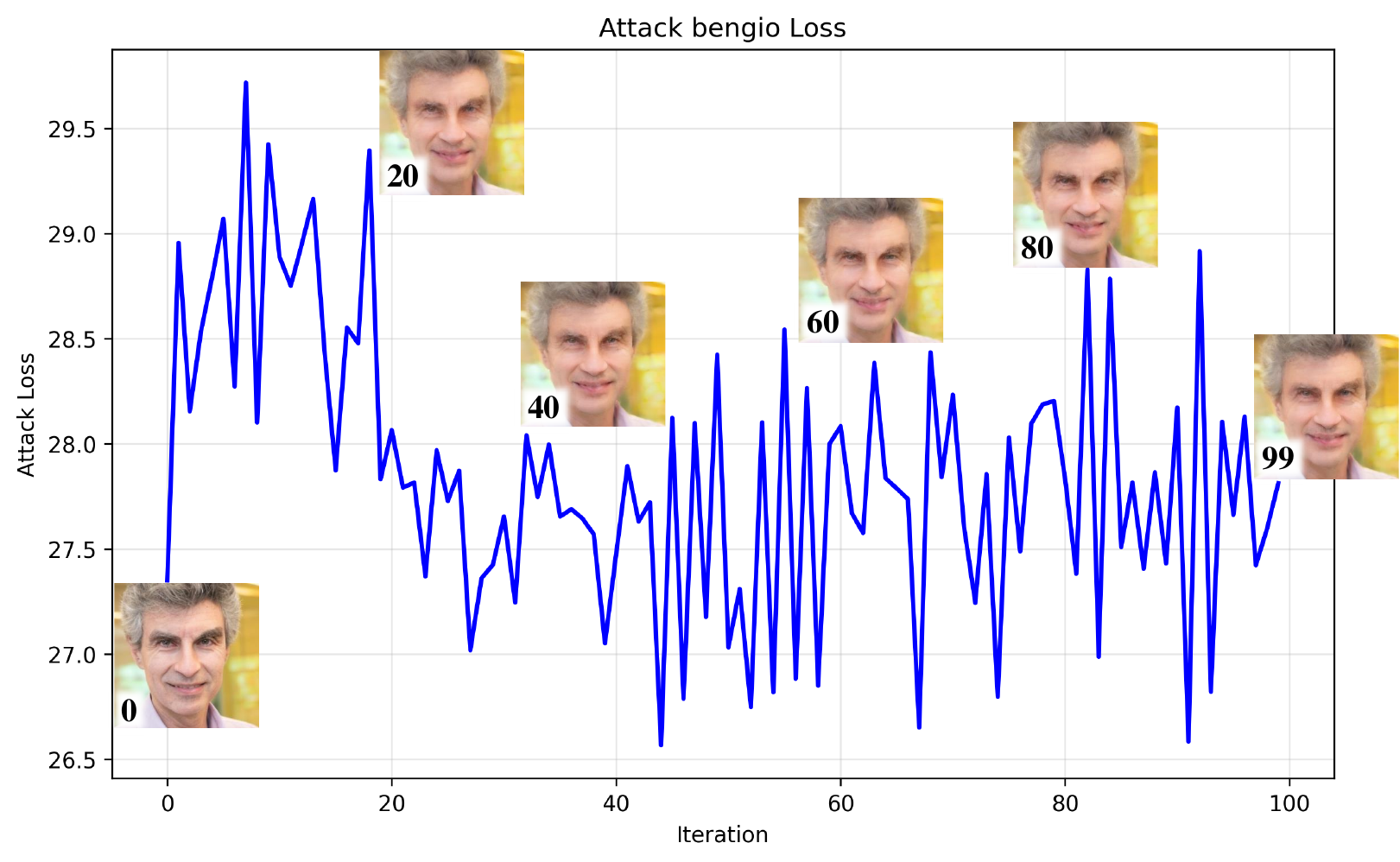}
    \caption{The $\mathcal{L}_{attack}$ curve for the image “bengio.jpg” under joint optimization. The number on the every image denotes the current iteration round. We extract the adversarial example at current round and convert it to RGB space for visualization.}
    \label{fig:appzd}
\end{figure}

The results show that under joint optimization, $\mathcal{L}_{attack}$ continually oscillates, indicating that the two components of the Min-max optimization are competing simultaneously, preventing the adversarial example from converging. Moreover, intermediate adversarial examples appear blurred during optimization, suggesting that joint optimization fails to accomplish our optimization objective. In contrast, under alternating optimization, $\mathcal{L}_{attack}$ steadily increases, allowing the adversarial example to converge to a Nash equilibrium point. The adversarial examples maintain imperceptibility throughout the optimization process, demonstrating that alternating optimization is well-suited to our task.

\section{Implementation Details}
\subsection{Datasets Configuration}
\label{app:Datasets}
The five datasets include image resolutions ranging from 512x512 to 3088x3087. For each dataset, we carefully choose 100 face images covering diverse ethnicities, ages, genders, skin tones, scenes, and accessories as the source images $I_{src}$ to be protected. Before generating adversarial examples, each image is cropped and resized to 512x512 resolution. To evaluate the defense capability of the adversarial examples, we randomly select 7 images from~\cite{han2024face} as $I_{tar}$, including celebrities of different ages, genders, and artistic portraits. 

\subsection{Parameters Configuration}
\label{app:para}
We employ the PGD attack~\cite{madry2017pgd} with $Q$ and $P$ set to 1000 and 100, respectively. For the parameter $N$, as shown in Algorithm~\ref{alg:minimax_alternating}, a larger $N$ implies more alternating optimization steps between $\mathbf{w}$ and $\hat{z}_{src}$, leading to adversarial examples with stronger defense effectiveness and better imperceptibility, albeit at increased computational cost. We therefore set $N = 2$. The parameter $M$ is typically set equal to the number of DDIM reverse denoising steps (e.g., 25). However,~\cite{yu2024step} demonstrates that using only 80\% of the original denoising steps can yield the best defense performance against diffusion models; we thus set $M = 20$. All experiments are conducted on an NVIDIA RTX 5090 GPU. The step size $\eta_z$ is set to 1/255, and $\eta_w$ is set to 0.001. The editing strength $\vartheta=0.95$ and the editing region control factor $\varrho=1.0$. The mapping $\delta \mapsto \boldsymbol{\delta}$ resamples $\delta$ to multi-scale noise tensors via adaptive-pooling (for resolutions $\leq 64\times64$) or bilinear interpolation (for higher resolutions), matching the StyleGAN2 generator’s noise inputs at each scale. For optimizing $\mathbf{w}$, we use the Adam optimizer~\cite{kingma2014adam}. Considering common user preferences, $z$ target attributes are selected as \texttt{'wearing\_lipstick'}, \texttt{'mouth\_slightly\_open'}, \texttt{'big\_nose'}, \texttt{'bushy\_eyebrows'}, and \texttt{'smiling'}. The mapping between the target attributes and the corresponding facial regions is as follows: \texttt{'wearing\_lipstick'} corresponds to the mouth; \texttt{'mouth\_slightly\_open'} corresponds to the mouth and teeth; \texttt{'big\_nose'} corresponds to the nose; \texttt{'bushy\_eyebrows'} corresponds to the eyebrows; and \texttt{'smiling'} corresponds to the mouth, eyebrows, and eyes. Users can also choose attributes corresponding to different facial regions based on their preferences. The hyperparameters balancing the loss terms are set as follows: $\lambda_1 = 19$, $\lambda_2 = 2.6$, $\lambda_3 = 0.13$, $\lambda_4 = 2$, $\lambda_5 = 0.005$, $\lambda_6 = 1$. We note that MaskFaceGAN~\cite{pernuvs2023maskfacegan} did not release code for multi-attribute editing and lacked a face alignment and cropping pipeline; we have reimplemented these functionalities accordingly. To improve computational efficiency during the backpropagation of $\mathcal{L}_{adv}$, we adopt an accelerated approximation strategy following the approach in~\cite{yam2025my}.

\section{Limitations and Potential Directions}
\label{appc:lim}
In this section, we discuss the limitations of the proposed method. On one hand, our approach necessitates alternating collaborative optimization of two distinct variables and involves multiple rounds of iterations, which incurs significant time overhead. For example, when $N=2$, generating a single adversarial example with our method requires approximately 5 minutes, hindering practical usability for users. Thus, a promising future direction is to accelerate the generation of adversarial examples for more efficient protection of users' identity information. On the other hand, our method relies on the specific attribute editing model (MaskFaceGAN), which is GAN-based and exhibits substantial architectural differences from DMs. This leads to a discrepancy between the latent spaces of the optimized variables $\mathbf{w}$ and $\hat{z}_{src}$. Additionally, MaskFaceGAN requires training and thus suffers from poor generalization, offering only a limited set of editable attributes, which restricts flexibility of our method. Consequently, another future research direction is to integrate flexible multi-attribute editing methods that share the same architecture, thereby ensuring that the optimized variables reside in a unified latent space and effectively reducing the min-max game problem to optimizing a single variable globally. Lastly, enhancing the robustness and transferability of our method is crucial to defend against adversaries who may employ more advanced face-swapping models.

\section{More Qualitative Comparisons}
\label{appd:visize_results}
\begin{figure*}[htbp]
    \centering
    \includegraphics[width=0.92\textwidth]{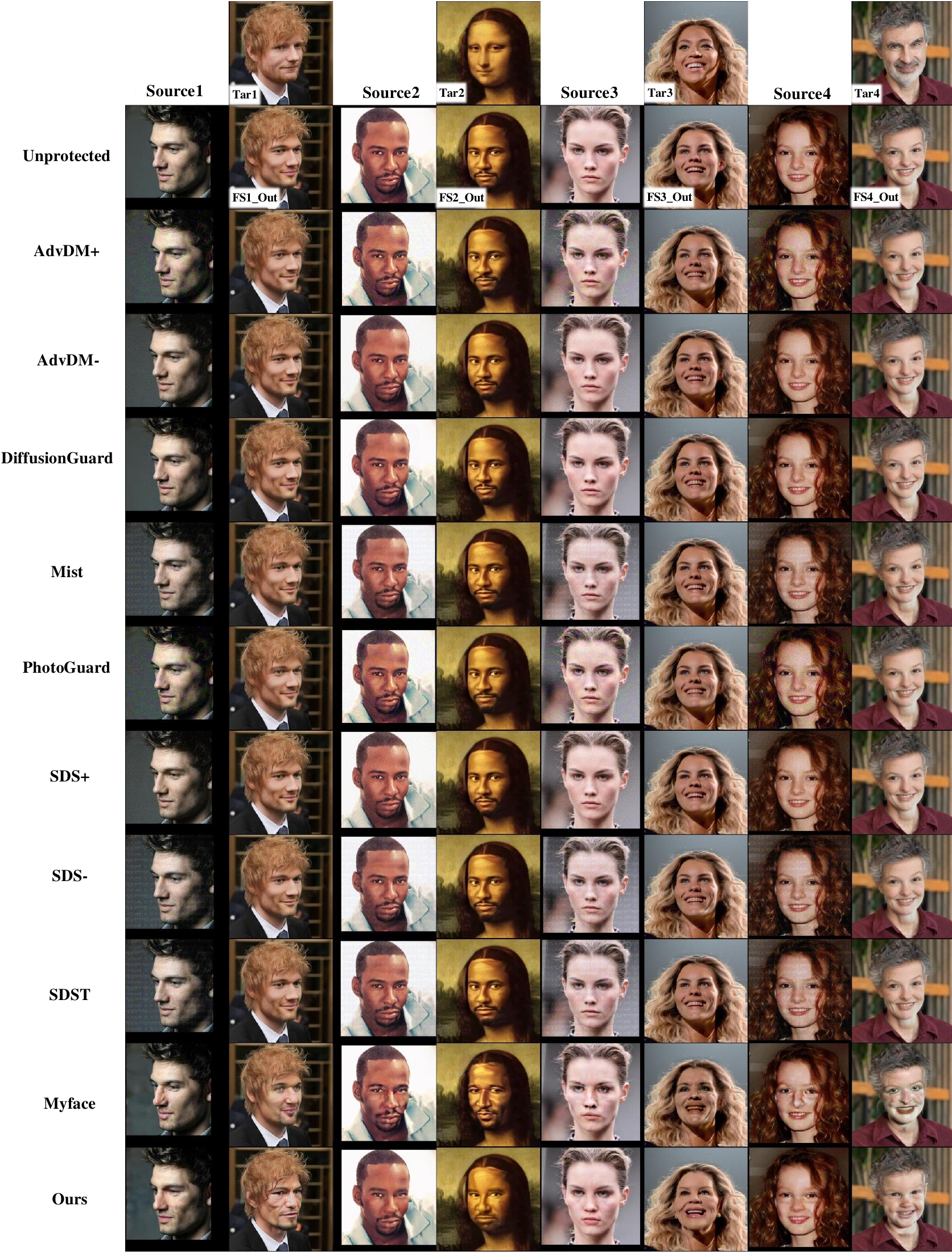}
    \caption{Qualitative comparison between other methods and our proposed approach. Here, MyFace and our method use $\epsilon=75/255$, while the $\epsilon$ values for other methods follow the optimal settings in their respective papers. The dataset is CelebA-HQ. }
    \label{qunoise-cele}
\end{figure*}

\begin{figure*}[htbp]
    \centering
    \includegraphics[width=0.92\textwidth]{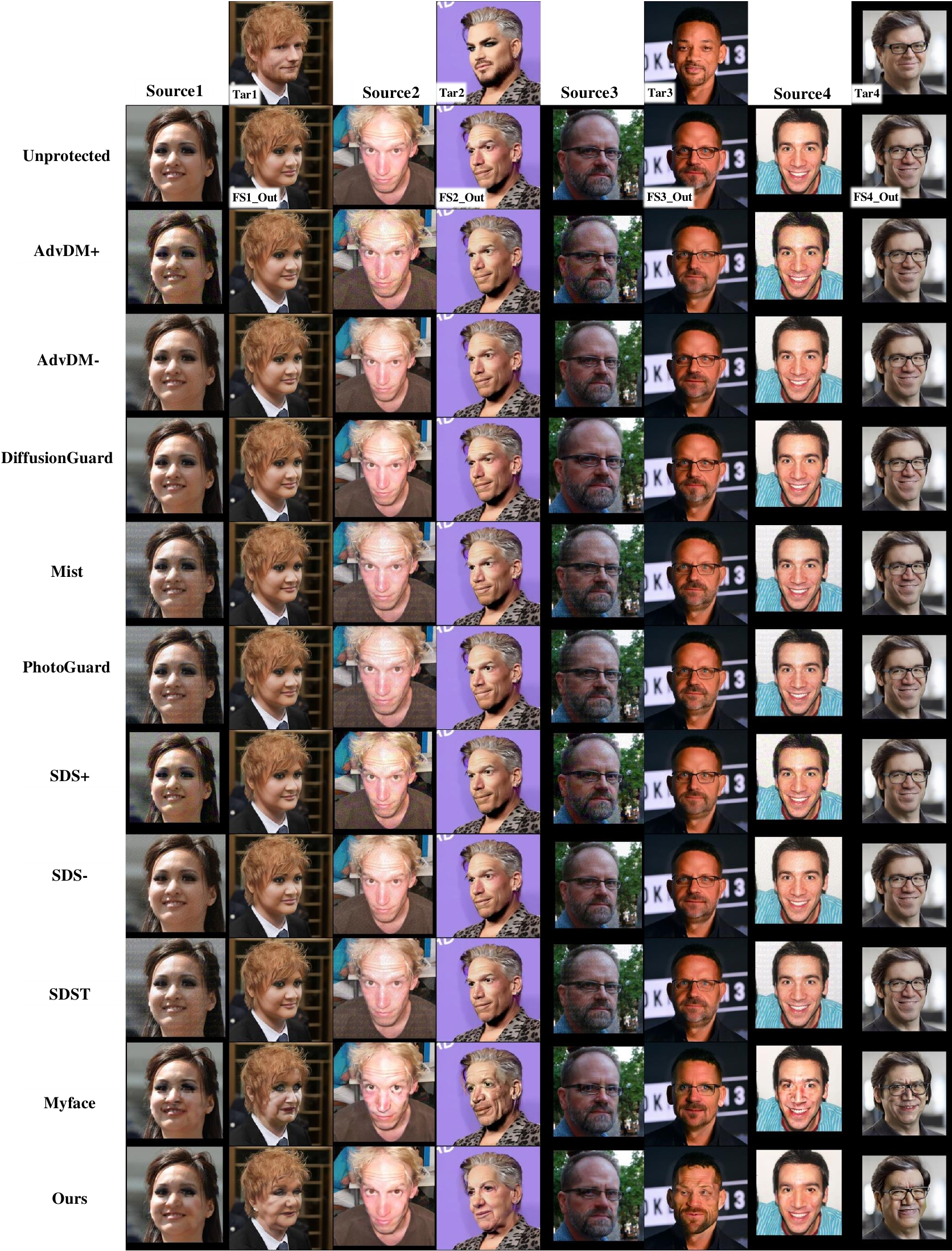}
    \caption{Qualitative comparison between other methods and our proposed approach. Here, MyFace and our method use $\epsilon=75/255$, while the $\epsilon$ values for other methods follow the optimal settings in their respective papers. The dataset is FFHQ. }
    \label{qunoise-ffhq}
\end{figure*}

\begin{figure*}[htbp]
    \centering
    \includegraphics[width=0.92\textwidth]{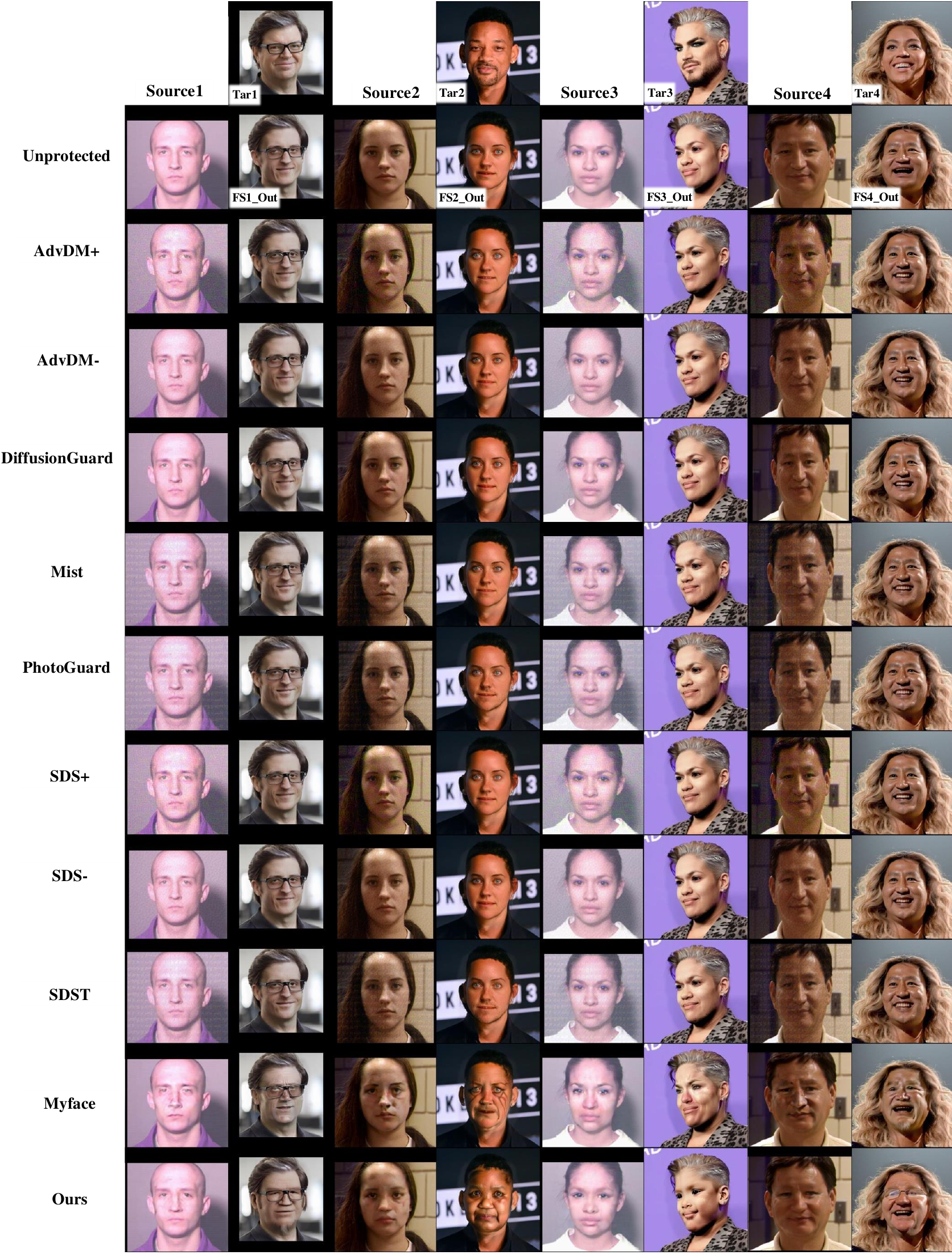}
    \caption{Qualitative comparison between other methods and our proposed approach. Here, MyFace and our method use $\epsilon=75/255$, while the $\epsilon$ values for other methods follow the optimal settings in their respective papers. The dataset is FRGC. }
    \label{qunoise-frgc}
\end{figure*}

\begin{figure*}[htbp]
    \centering
    \includegraphics[width=0.92\textwidth]{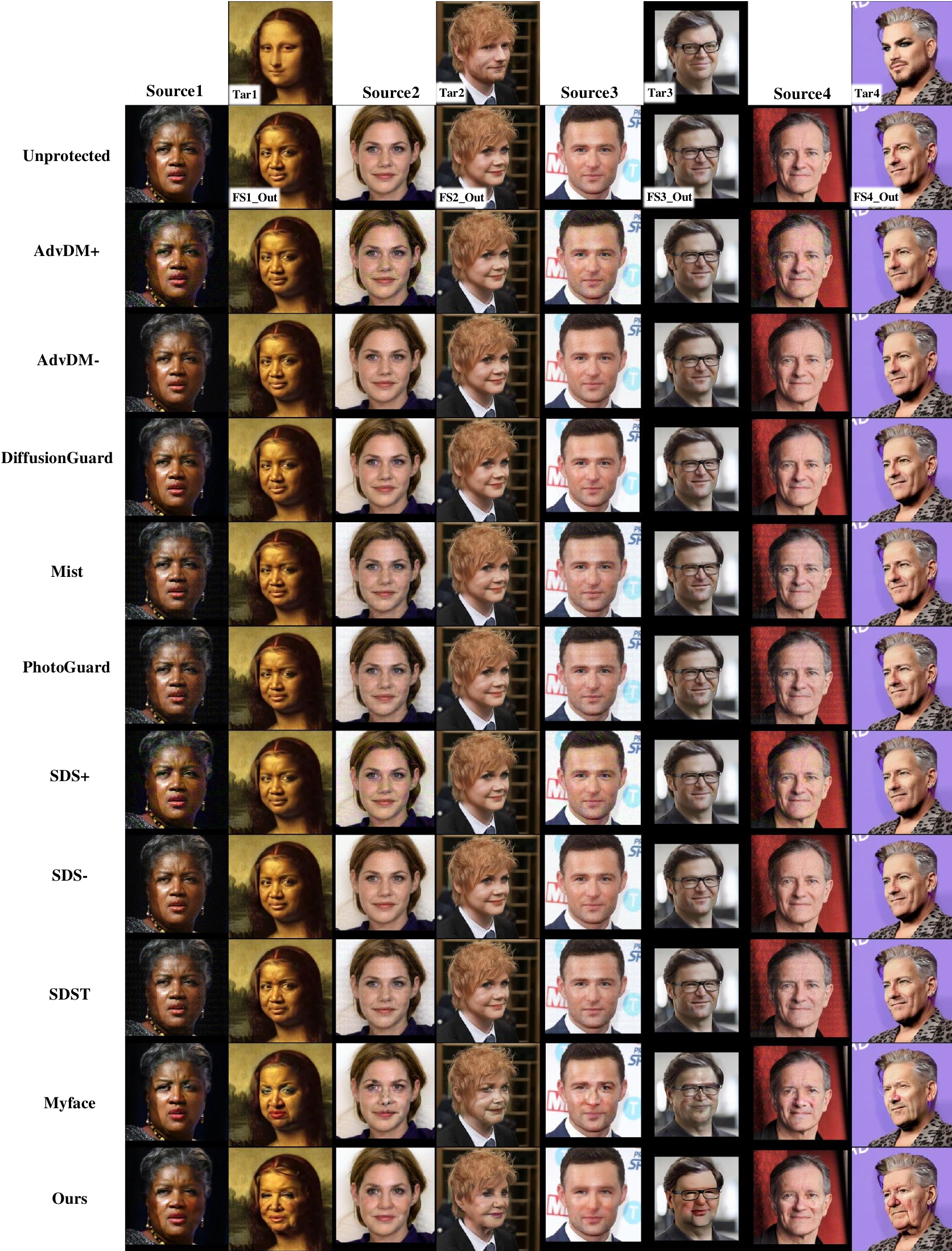}
    \caption{Qualitative comparison between other methods and our proposed approach. Here, MyFace and our method use $\epsilon=75/255$, while the $\epsilon$ values for other methods follow the optimal settings in their respective papers. The dataset is VoxCeleb2. }
    \label{qunoise-vox}
\end{figure*}

\begin{figure*}[htbp]
    \centering
    \includegraphics[width=0.92\textwidth]{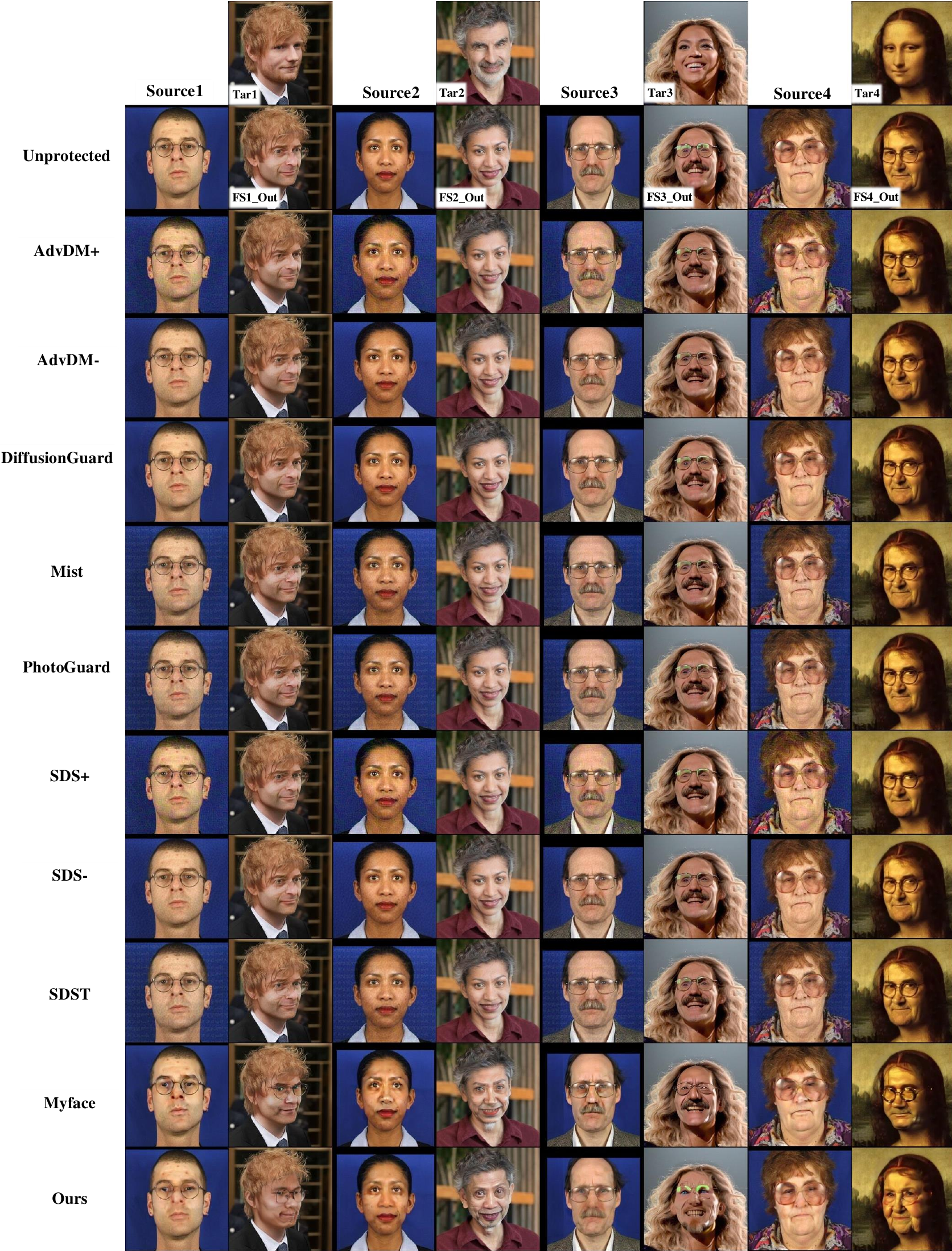}
    \caption{Qualitative comparison between other methods and our proposed approach. Here, MyFace and our method use $\epsilon=75/255$, while the $\epsilon$ values for other methods follow the optimal settings in their respective papers. The dataset is XM2VTS. }
    \label{qunoise-xm2}
\end{figure*}

\begin{figure*}[htbp]
    \centering
    \includegraphics[width=1.02\textwidth]{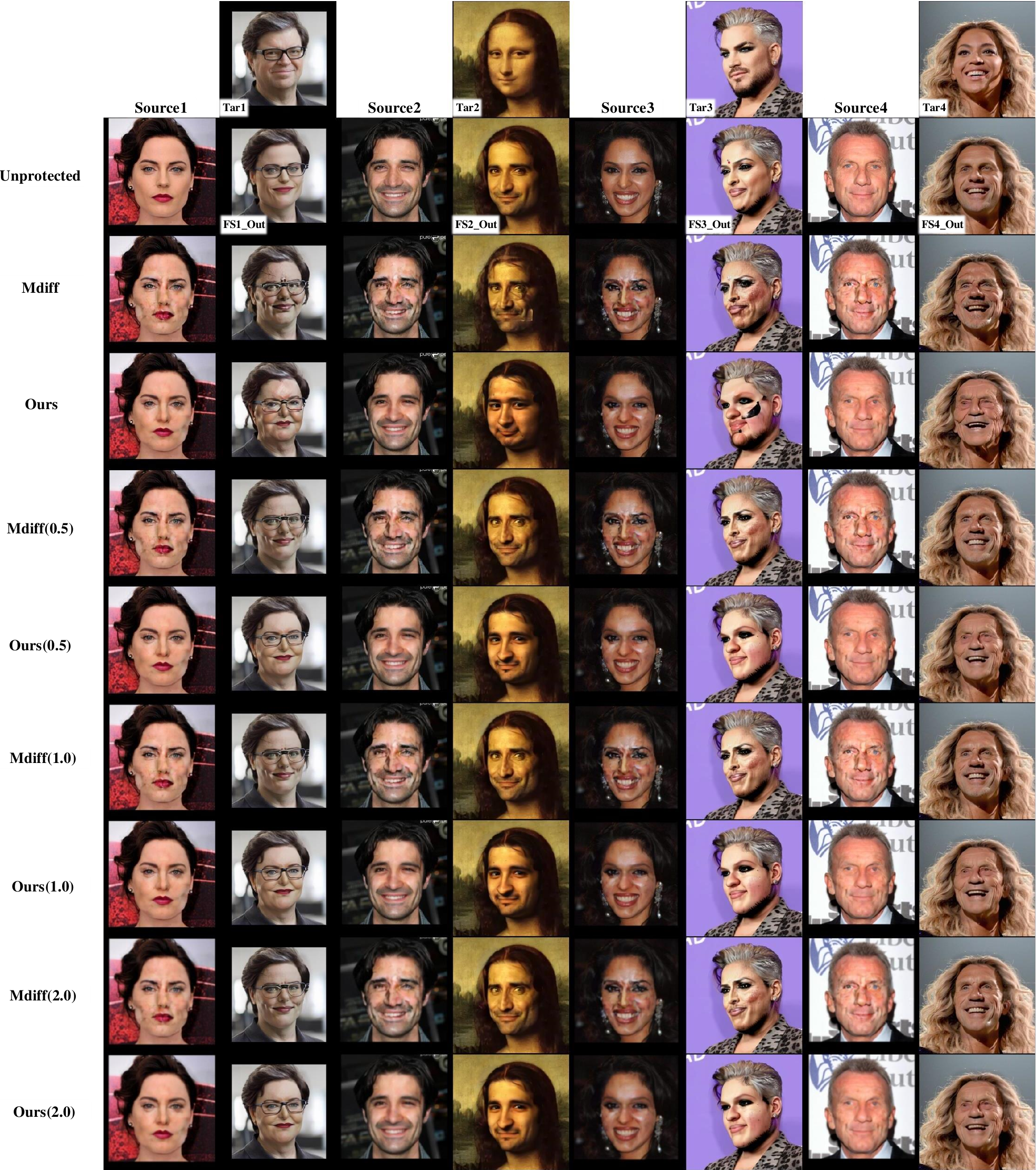}
    \caption{Additional qualitative comparison between Mdiff~\cite{yam2025my} and our proposed method under Gaussian blur. Here, $\epsilon=75/255$, and the dataset is CelebA-HQ. The values in parentheses denote the $\sigma$ parameter. }
    \label{qurobust-guess}
\end{figure*}

\begin{figure*}[htbp]
    \centering
    \includegraphics[width=1.02\textwidth]{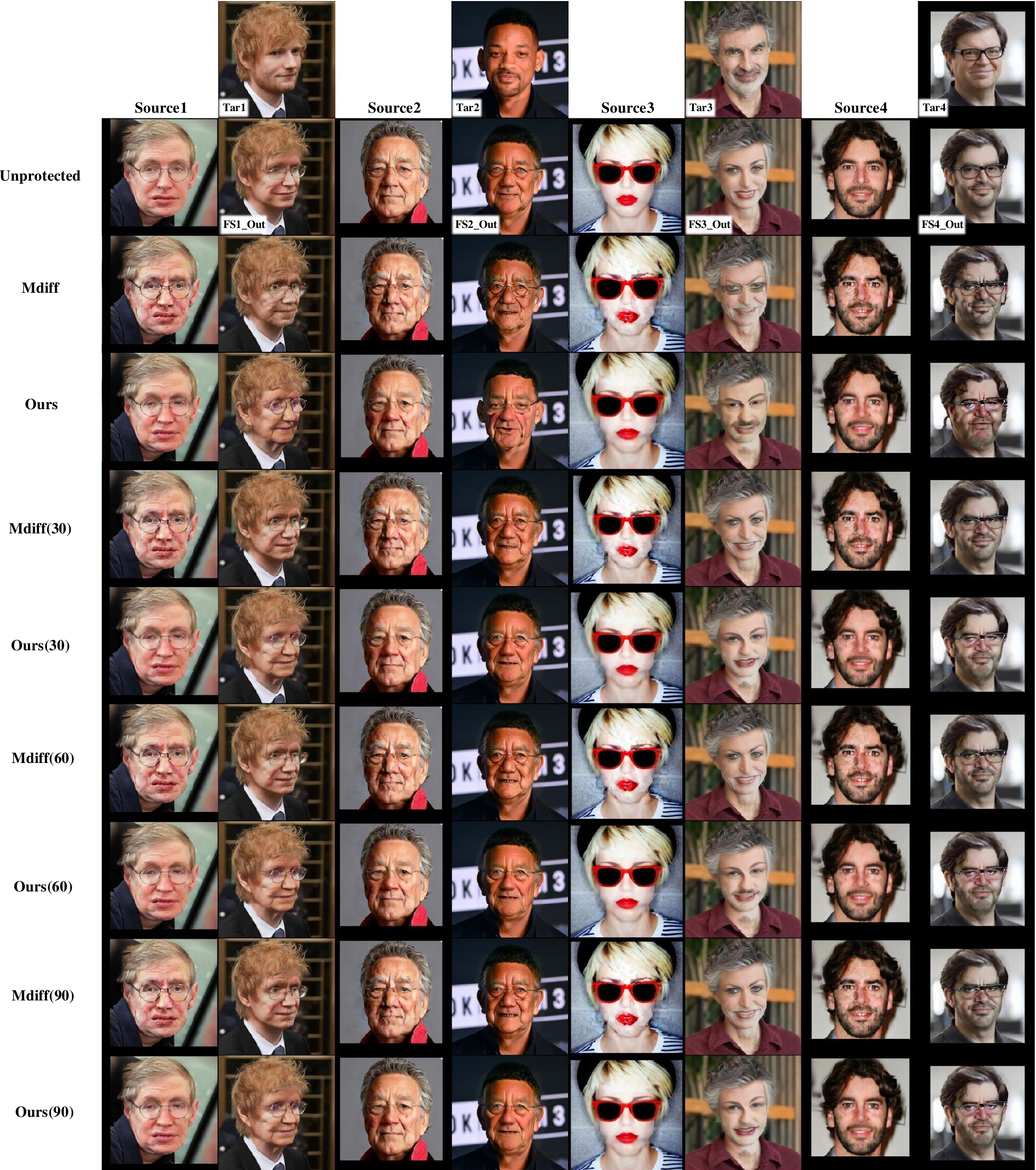}
    \caption{Additional qualitative comparison between Mdiff~\cite{yam2025my} and our proposed method under JPEG compression. Here, $\epsilon=75/255$, and the dataset is CelebA-HQ. The values in parentheses denote the $q$ parameter. }
    \label{qurobust-jpeg}
\end{figure*}

\begin{figure*}[htbp]
    \centering
    \includegraphics[width=0.96\textwidth]{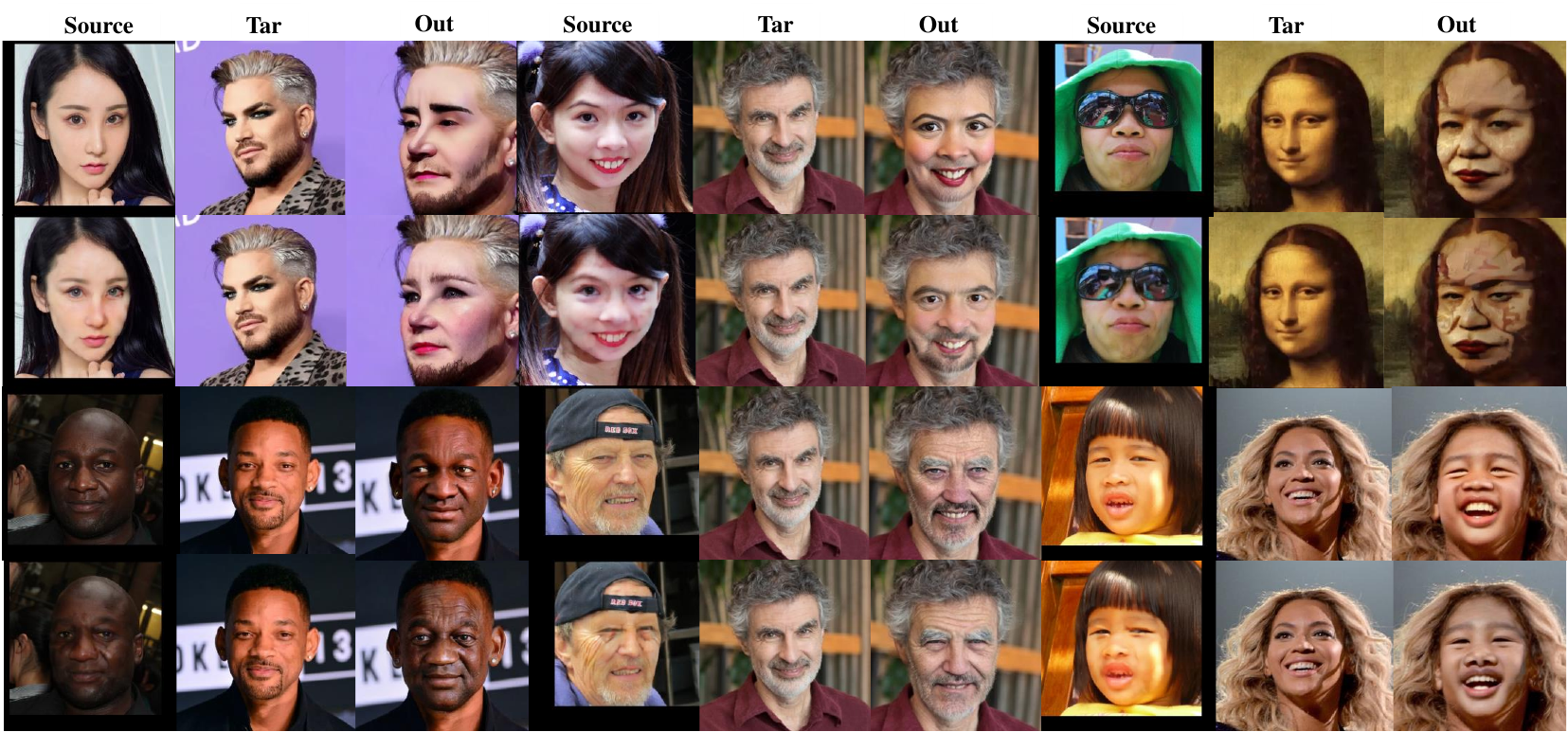}
    \caption{Qualitative results of our proposed method defending the REFace~\cite{baliah2024} model. Here, $\epsilon=75/255$, and the dataset is FFHQ. The first and third rows show face-swapping results using clean samples, while the second and fourth rows show the results after using adversarial examples. }
    \label{trans-reface}
\end{figure*}

\begin{figure*}[htbp]
    \centering
    \includegraphics[width=0.92\textwidth]{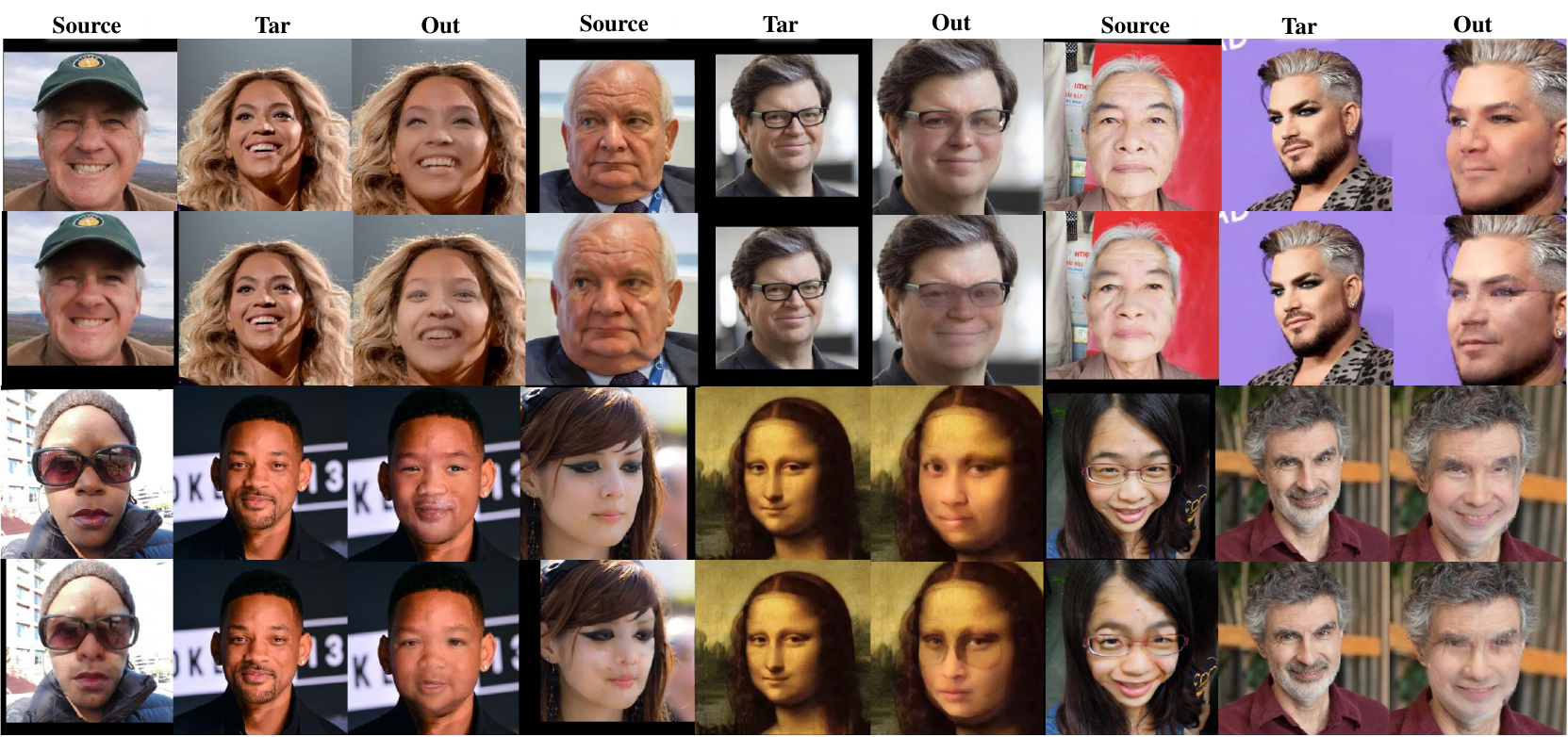}
    \caption{Qualitative results of our proposed method defending the DiffSwap~\cite{zhaoDiffSwap2023a} model. Here, $\epsilon=75/255$, and the dataset is FFHQ. The first and third rows show face-swapping results using clean samples, while the second and fourth rows show the results after using adversarial examples. }
    \label{trans-diffswap}
\end{figure*}

\begin{figure*}[htbp]
    \centering
    \includegraphics[width=0.90\textwidth]{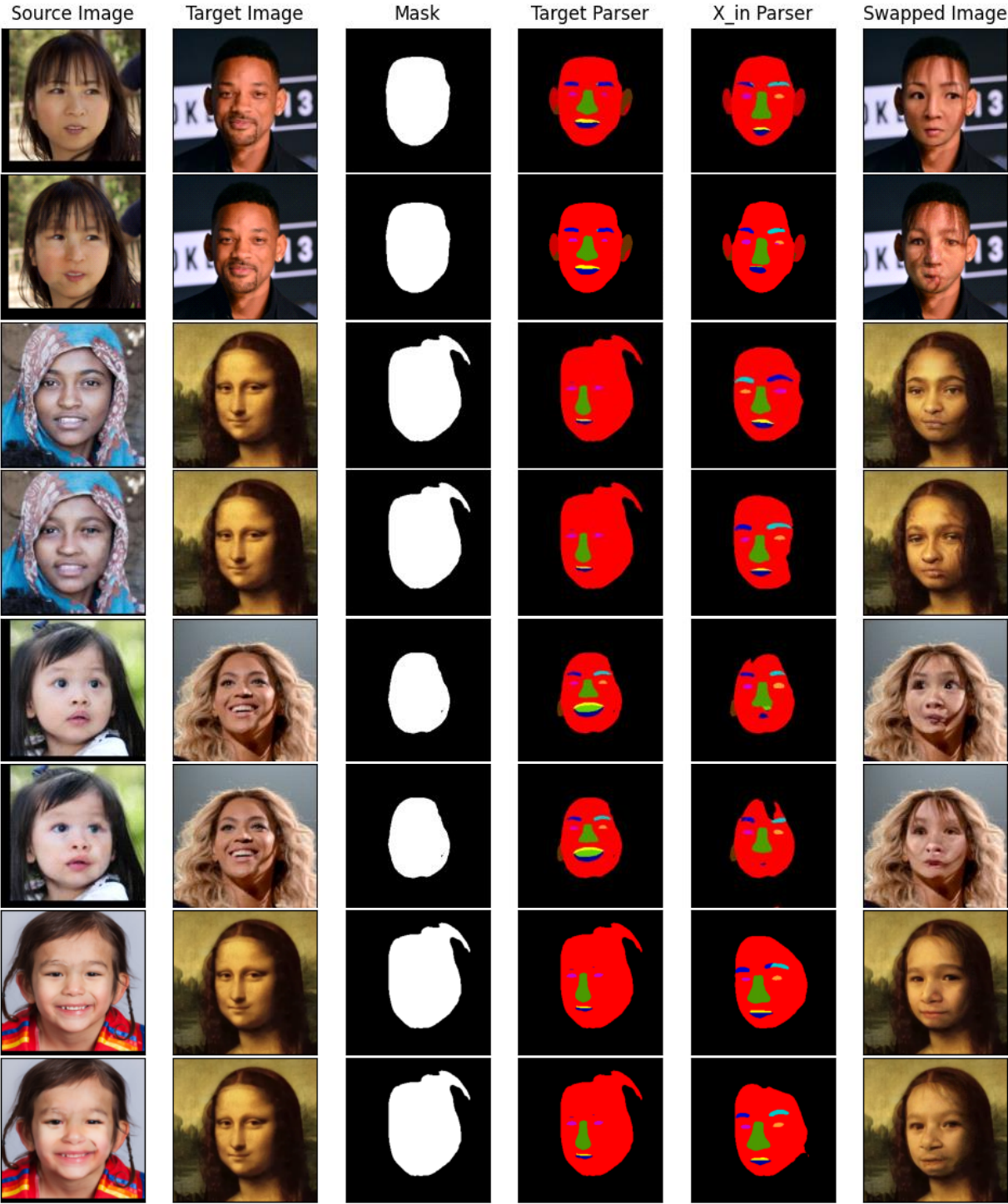}
    \caption{Qualitative results of our proposed method defending the DiffFace~\cite{kimFace2025} model. Here, $\epsilon=75/255$, and the dataset is FFHQ. The first, third, fifth and seventh rows show face-swapping results using clean samples, while the second, fourth, sixth and eighth rows show the results after using adversarial examples. }
    \label{trans-diffface}
\end{figure*}

In this section, we provide additional qualitative comparisons for readers to evaluate. As shown in Figure~\ref{qunoise-cele}--\ref{trans-diffface}.

\end{document}